\documentclass[10pt,twocolumn,letterpaper]{article}

\usepackage{cvpr}              %

\usepackage[dvipsnames]{xcolor}

\definecolor{cvprblue}{rgb}{0.21,0.49,0.74}
\usepackage[pagebackref,breaklinks,colorlinks,citecolor=cvprblue]{hyperref}
\usepackage{overpic}
\usepackage{multirow}
\usepackage{multicol}
\usepackage{adjustbox}
\usepackage{scalerel}

\title{VXP: Voxel-Cross-Pixel Large-Scale Camera-LiDAR Place Recognition}

\author{Yun-Jin Li$^{1,2\star}$
\and
Mariia Gladkova$^{1,2\star}$
\and
Yan Xia$^{1,2\dagger}$
\and
Rui Wang$^{3}$
\and
Daniel Cremers$^{1,2}$\\
{\centering $^1$ TU Munich $^2$ Munich Center for Machine Learning $^3$ Microsoft}\\
{\tt \centering \small \{yunjin.li, mariia.gladkova, yan.xia, cremers\}.@tum.de \hspace{1cm} wangr@microsoft.com}
}
\begin{document}
\maketitle
\let\oldthefootnote\thefootnote
\renewcommand{\thefootnote}{\fnsymbol{footnote}} 
\footnotetext[2]{Corresponding author. * Equal contribution.} 
\let\thefootnote\oldthefootnote
\begin{abstract}
 Cross-modal place recognition methods are flexible GPS-alternatives under varying environment conditions and sensor setups. However, this task is non-trivial since extracting consistent and robust global descriptors from different modalities is challenging. To tackle this issue, we propose \textit{Voxel-Cross-Pixel (VXP)}, a novel camera-to-LiDAR place recognition framework that enforces local similarities in a self-supervised manner and effectively brings global context from images and LiDAR scans into a shared feature space. Specifically, VXP is trained in three stages: first, we deploy a visual transformer to compactly represent input images. Secondly, we establish local correspondences between image-based and point cloud-based feature spaces using our novel geometric alignment module. We then aggregate local similarities into an expressive shared latent space. Extensive experiments on the three benchmarks (Oxford RobotCar, ViViD++ and KITTI) demonstrate that our method surpasses the state-of-the-art cross-modal retrieval by a large margin. Our evaluations show that the proposed method is accurate, efficient and light-weight. 
Our project page is available at: \href{https://yunjinli.github.io/projects-vxp/}{https://yunjinli.github.io/projects-vxp/}.
\end{abstract}

\section{Introduction}
\label{sec:intro}
Since the emergence of autonomous systems, global place recognition has become essential for mobile robotics. Despite the widespread availability of the Global Navigation Satellite System (GNSS), signal outages remain inevitable, particularly in parking spaces or urban areas where buildings or tunnels can block satellite signals~\cite{wen2019toward}. These disruptions are critical challenges for achieving autonomous driving on a city-wide scale and must be managed using onboard devices like cameras~\cite{arandjelovic2016netvlad}, LiDARs~\cite{xia2023lightweight}, or radars~\cite{wang2021rodnet}. 
The Autonomous Vehicle (AV) sensor suite provides various strategies for data recording and, thus, enables alternative ways for global localization in GNSS-denied areas. 
Although numerous solutions have been proposed within the computer vision and robotics communities, most still rely on the same type of data during both map acquisition and operation. This dependence on a single data source may limit the applicability of these solutions in cases of sensor malfunctions or variations in sensor setups.
Consequently, there is a need for more flexible localization methods that can take advantage of different sensor modalities under varying environmental conditions. This presents significant potential for cross-modal place recognition techniques. While multi-modal approaches require data to be available from all sensors, cross-modal methods are intended to be more flexible and seamlessly switch between the map and query sources. For instance, camera-to-LiDAR method would support querying a database of encoded LiDAR scans with RGB images (2D-3D localization). In terms of practical value, it would save the on-board computational load of processing large point clouds and guarantee global localization even in cases of LiDAR malfunctioning using image data.\\
Although cross-modal place recognition offers significant potential, it also presents challenges due to substantial differences between observations from various sensors. Specifically, in camera-to-LiDAR localization, images and point clouds exhibit a clear gap in both raw data (2D images vs 3D scans) and extracted features. The lack of explicit correlation between these two data modalities complicates the development of cross-modal global localization solutions. 
Due to this, only a few approaches have been proposed to tackle the task so far. 
Cattaneo et al. \cite{cattaneo2020global} first introduce 2D and 3D feature extraction networks to create a shared embedding space between images and point clouds.  $LC^2$~\cite{lee20232} proposes to transform image and point clouds into the same 2.5D space for reducing the domain gap. LIP-Loc~\cite{shubodh2024lip} advocates usage of multi-class N-pair batched loss in the contrastive learning regime to boost cross-modal retrieval. While these methods focus on designing powerful networks to encode data into robust global descriptors, they ignore geometric relation between local structures captured by both modalities. Local consistency not only provides additional constraints in order to effectively bridge the domain gap during the cross-modal training, but also enhances the representative power of the shared latent space.

\begin{figure*}
  \centering
  \begin{subfigure}{0.48\linewidth}
    \includegraphics[width=1\linewidth]{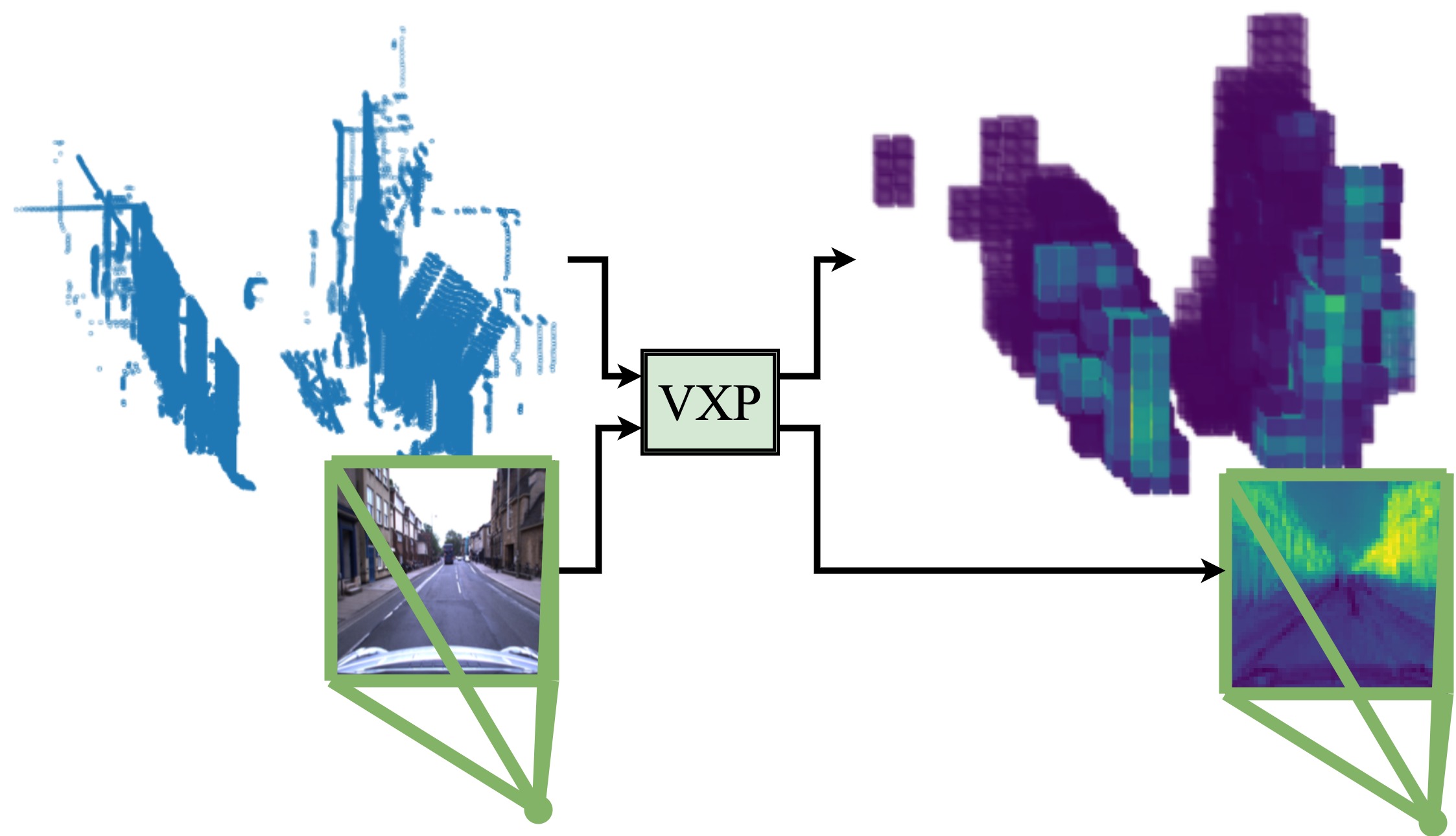}
    \end{subfigure}
\begin{subfigure}{0.48\linewidth}
    \includegraphics[width=1\linewidth]{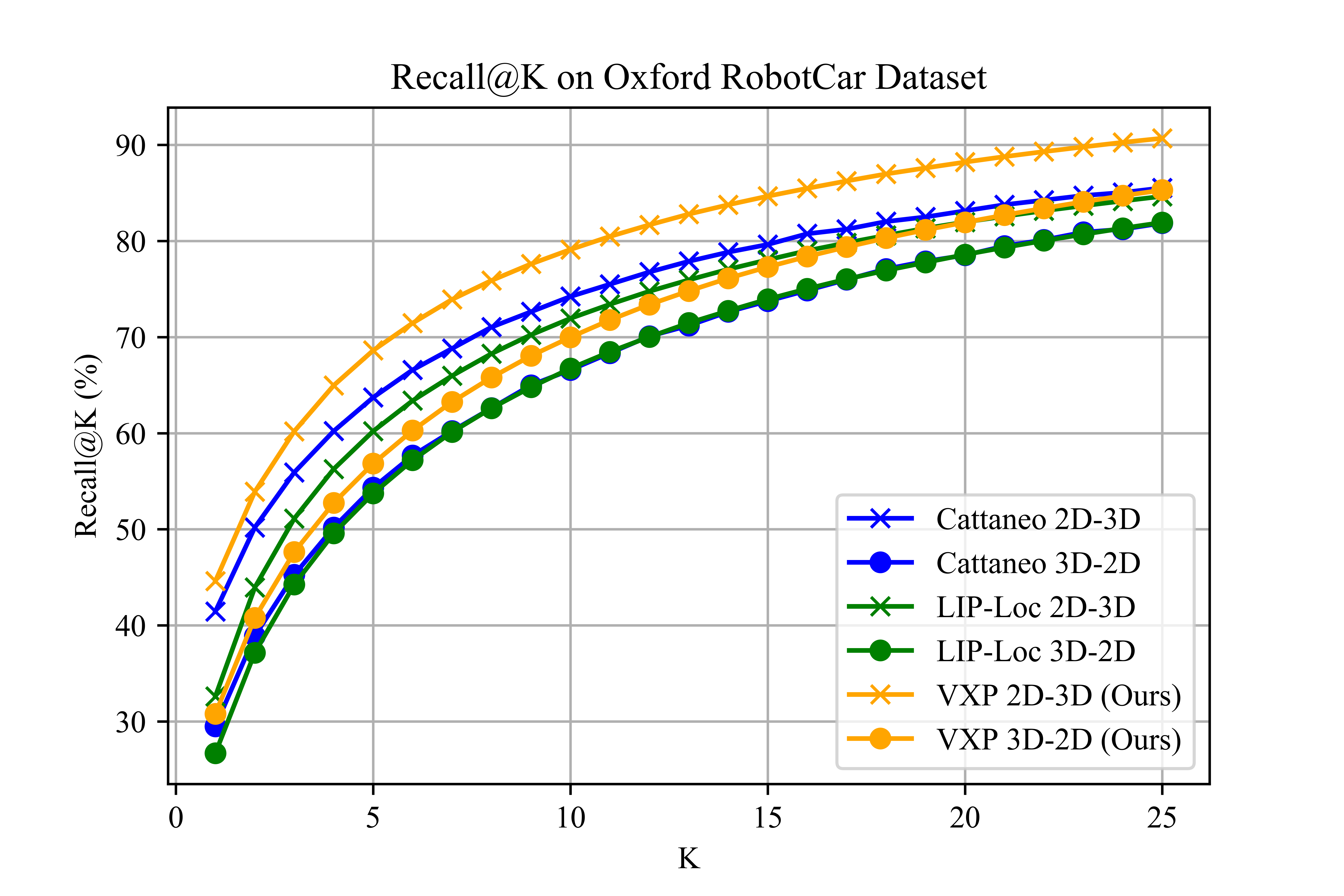}
    \end{subfigure}
\caption{(\textbf{Left}) Voxel-Cross-Pixel (VXP) can effectively map data from different modalities (2D images and 3D LiDAR scans) into the shared latent space, which exhibits local similarities and captures global context. (\textbf{Right}) Recall for up-to K = 25 retrieved places on Oxford RobotCar benchmark. VXP consistently demonstrates superior cross-modal large-scale global retrieval preformance.}
  \label{fig:teaser_fig}
\end{figure*}
\noindent In light of this, we introduce a novel method {\em Voxel-Cross-Pixel (VXP)} for camera-LiDAR place recognition. Our pipeline is three-fold: firstly, we leverage the power of visual transformers to obtain an expressive feature map and compact global embedding for an input image. Secondly, we choose sparse voxelized representation of a corresponding LiDAR scan and hierarchically aggregate features by utilizing sparse 3D convolutions. By means of projective geometry we establish local feature correspondences between image- and voxel-based feature maps and enforce their similarity during training. Lastly, we enforce similarities between global descriptors of the cross-modal matches. This comprehensive training paradigm enables the network to effectively capture both fine-grained local details and broader global context, facilitating successful cross-modal learning. We evaluate our model on three real-world datasets, achieving state-of-the-art cross-modal retrieval.\\
\noindent To summarize, the main contributions of the paper are:
\begin{itemize}
    \item We propose a novel framework for the cross-modal place recognition, \textit{Voxel-Cross-Pixel (VXP)}, which effectively encodes images and LiDAR scans in a shared latent space.
    \item We demonstrate the effectiveness of local similarity constraints in learning robust global descriptors for the cross-modal place recognition task.
    \item We establish state-of-the-art performance in cross-modal retrieval on the Oxford RobotCar, ViViD++ datasets and KITTI benchmark, while maintaining high uni-modal global localization accuracy.
    \item We publicly release our code along with implemented baselines at: \href{https://github.com/yunjinli/vxp}{https://github.com/yunjinli/vxp}.
\end{itemize}

\section{Related Work}
In this section, we first review uni-modal place recognition techniques. We then introduce some fusion-based approaches. Finally, the existing cross-modal methods are presented.\\
\noindent \textbf{Visual and point cloud-based retrieval.}
Uni-modal place recognition methods operate within one sensor type and aim to find the closest query match in a database. Most widely researched modalities are visual and LiDAR-based, while other types such as radar recently have received attention from the community~\cite{RadarRobotCarDatasetArXiv}.
Traditional image-based approaches, such as bag-of-words \cite{galvez2012bags}, represent different places with a visual vocabulary of quantized local descriptors~\cite{philbin2007object} and they are widely used in the SLAM community for re-localization and loop closure tasks~\cite{campos2021orb, gladkova2021tight}. In recent years, Convolutional Neural Network (CNN)-based methods have gained popularity for their expressiveness and enhanced robustness. Arandjelovi{\'c} et al. introduced NetVLAD \cite{arandjelovic2016netvlad}, a CNN-based approach that encodes RGB images into dense feature maps and learns to effectively aggregate these features into a global descriptor. CosPlace~\cite{berton2022rethinking} explored to perform the retrieval as a
classification task. Recent works~\cite{ali2022gsv, ali2023mixvpr} proposed to process the features extracted by a CNN with a Conv-AP layer or a Feature-Mixer. AnyLoc \cite{keetha2023anyloc} utilizes the features generated from off-the-shelf self-supervised model (DINOv2 \cite{oquab2023dinov2}) to achieve SOTA performance in many VPR benchmarks.\\
As for LiDAR-based place recognition, Uy et al. proposed PointNetVLAD \cite{uy2018pointnetvlad}, in which they employed PointNet \cite{qi2017pointnet} to extract features from a point cloud map and then aggregate them into a global descriptor using a subsequent NetVLAD layer. LPD-Net was introduced by Liu et al. \cite{liu2019lpd}, in which an adaptive local feature extraction module is proposed to extract local features along with the graph-based aggregation module to effectively combine them. SOE-Net~\cite{xia2021soe} first introduces orientation encoding into PointNet and a self-attention unit to generate a robust 3D global  descriptor. Furthermore, various methods~\cite{zhou2021ndt, fan2022svt} explored the integration of different transformer networks to learn long-range contextual relationships.
In contrast, Minkloc3D~\cite{komorowski2021minkloc3d} employed a voxel-based strategy to generate a compact global descriptor. However, the voxelization methods inevitably suffer from information loss due to the quantization. Recent CASSPR~\cite{Xia_2023_ICCV} thus introduced a hierarchical cross attention transformer, combining both the advantages of voxel-based strategies with the point-based strategies. Text2Loc~\cite{xia2024text2loc} achieved the 3D localization based on textual descriptions. In this paper, our work brings the best practices of 2D image and 3D point cloud communities together into a coherent framework that can achieve state-of-the-art performance in cross-modal retrieval.\\
\noindent \textbf{Fused-Modal Place Recognition.}
LiDAR-based methods are more robust to variations in illumination and appearance when compared to the vision-based approaches. However, obtained scans are limited in capturing fine details of the observed scenes, while image data offers rich and dense scene capture. To this end, researchers have started exploring the possibility of fusing image and LiDAR data for the place recognition task. Pan et al. proposed a method called CORAL \cite{pan2021coral}, in which point cloud data is converted into an elevation image in order to perform further fusion. MinkLoc++ \cite{komorowski2021minkloc++}, on the other hand, employed a late fusion technique, processing point cloud and image data separately and performing fusion at the final stage. While our approach relies on having both image and LiDAR data available during training, due to the chosen architecture with two independent branches we are capable of dealing with a single stream data during inference, which enables cross-modal retrieval.\\  
\noindent \textbf{Cross-Modal Place Recognition.}
Cattaneo et al. \cite{cattaneo2019cmrnet} were the first to introduce this task, proposing a data-driven method where two networks were trained to encode images and point cloud maps separately using a teacher-student training approach. Initially, the image network (teacher) was trained using the triplet loss function \cite{schroff2015facenet}, and then the point cloud network (student) was trained to align point embeddings within the shared latent space. In our work, we build on this paradigm with a stronger image backbone and enhance global descriptors by incorporating local feature constraints. The $LC^2$ approach, proposed by Lee et al.~\cite{lee20232}, presented an alternative method for cross-modal retrieval, where the domain gap was bridged by pre-processing sensor data and transforming it into the same data representations. Specifically, they converted both types of data into the 2.5D space, where RGB images were turned into disparity maps using depth network~\cite{watson2021temporal} and LiDAR point clouds were transformed into range images. A self-supervised pre-training scheme \cite{leyva2023data} was employed on the encoders, enabling the networks convergence. Similar method such as Lip-Loc~\cite{shubodh2024lip} also proposed to process LiDAR-scans into range images and optimize their encoders by contrastive learning. In comparison, our method directly handles input raw data and does not require computationally demanding pre-processing steps such as generation of range images or depth maps, which would be more favorable for on-board devices.\\
A few studies have been proposed to tackle cross-modal registration such as 2D-3D re-localization~\cite{li2021deepi2p, ren2022corri2p, wang2023end, feng20192d3d}. These methods primarily concentrate on accurately aligning a given camera view with a corresponding point cloud map and estimating relative 6-DoF transformation between them. In our work, we propose a solution for finding the cross-modal pairs, which are often unavailable in a real-world scenario, and advocate usefulness of local constraints in achieving this goal.

\section{Problem Statement}
We begin by defining the task of cross-modal place recognition. In particular, we are interested in camera-to-LiDAR retrieval, however the definition can be naturally extended to other modalities such as radars.\\
Given a reference map $M_\text{ref}$, where each element (a 2D image $I$ or a 3D point cloud P) is tagged with a GPS coordinate, we aim to retrieve the geographically closest match to a query $Q$ from a different sensor modality, such as LiDAR scanner or camera respectively. With this, the cross-modal place recognition can be defined formally for 3D-2D as $$I^* = \text{argmin} \{d(g(Q), f(I))\}$$ or for 2D-3D as
$$P^* = \text{argmin} \{d(f(Q), g(P))\},$$ where $d(\cdot)$ is a distance metric (e.g. L1 norm), $f$ is an image network, $g$ is a point cloud network and $I, P \in M_\text{ref}$. This step can be efficiently done using a KD-tree (e.g. from FAISS library~\cite{johnson2019billion}).
\section{Method}
\label{sec:vxp}
\begin{figure*}
  \centering
  \includegraphics[width=1\linewidth]{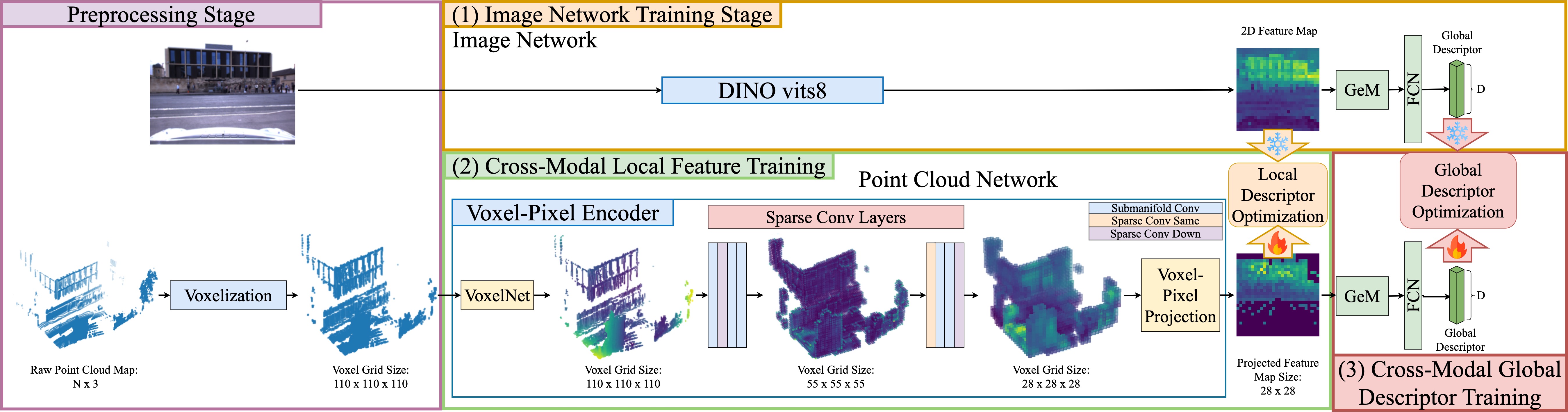}
  \caption{VXP pipeline comprises three steps:  (1) image network training (\cref{subsec:image_pretrain}), (2) cross-modal local feature training (\cref{subsec:local_train}), and cross-modal global descriptor training (\cref{subsec:global_train}). Starting from step (2) image features are frozen (~\protect\scalerel*{\includegraphics{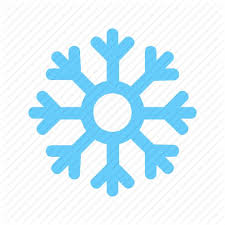}}{\strut}~), while the point cloud features are trained (~\protect\scalerel*{\includegraphics{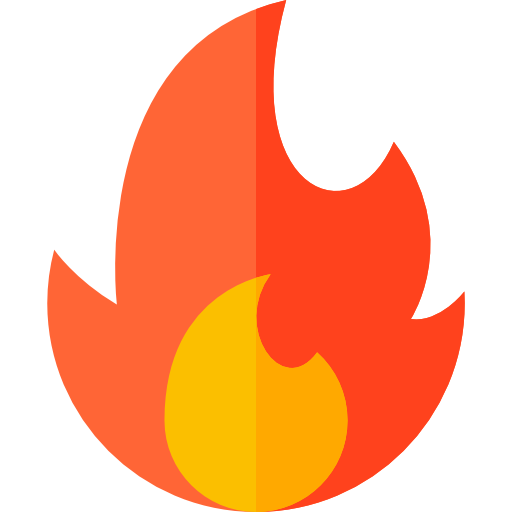}}{\strut}~). The two networks operate independently during inference, so queries and database samples can be processed separately. The objective is to map different data into a shared latent space and minimize the distance (e.g. L2 norm) between global descriptors of different modalities taken from the same space. }
  \label{fig:overview}
\end{figure*}
In this section, we introduce our cross-modal place recognition approach in detail. We design two separate networks that map image and point cloud into the shared latent space.\\
Practically, dealing with raw point cloud data, which typically consists of thousands of points, can pose a significant computational challenge. To tackle this problem, we downsample each input scan before feeding it to a network. To this end, we leverage point cloud grouping techniques, which has also been shown to effectively capture local structures~\cite{qi2017pointnet++}. Consequently, we deploy voxelization method~\cite{zhou2018voxelnet} to transform the raw point cloud data $\textbf{P}\in \mathbb{R}^{N \times 3}$ into a voxel grid $\textbf{V}=\{\textbf{v}_i\in \mathbb{R}^{M \times 3}, \textbf{c}_i \in \mathbb{R}^{3}\}_{1, 2, ..., T}$, where $T$ is the number of non-empty voxels and $M$ represents the maximal number of points within a voxel. If the number of points in a voxel is lower than $M$, we do zero-padding. From this point, the framework employs a voxel-based representation of LiDAR scans.\\
Our Voxel-Cross-Pixel (VXP) pipeline comprises three steps as demonstrated in \cref{fig:overview}. Firstly, we train an image network to learn distinctive global descriptors based on positive and negative image pairs (\cref{subsec:image_pretrain}). The learned feature space guides optimization in the second stage, where we enforce local correspondences by deploying the Voxel-Pixel Projection module in the point cloud branch (\cref{subsec:local_train}). Lastly, we optimize for the similarity between global descriptors to ensure consistency (\cref{subsec:global_train}).
\subsection{Image Network}
\label{subsec:image_pretrain}
The image network architecture comprises two components: (1) the DINO ViTs-8 encoder and (2) a global pooling layer (GeM + FCN) as illustrated in \cref{fig:overview}. In the initial phase, an RGB image, denoted as $I \in R^{ H\times W\times 3}$, is processed by the DINO ViTs-8 encoder $f^{enc}: R^{H\times W \times 3} \rightarrow R^{H^*\times W^* \times D}$, where $H^* = H // 8$ and $W^* = W // 8$. This operation yields 2D features, which are also recognized as local image feature descriptors. Subsequently, these generated image features are passed through the global pooling layer $f^{pool}: R^{H^*\times W^*\times D} \rightarrow R^{D}$, resulting in the creation of a global image descriptor.\\
\noindent We train the image network in a contrastive learning regime using a triplet loss function as per~\cref{eq:image_train}, where an anchor image denoted as $I^a_i$, a positive image $I^p_i$ closely related to the anchor image's location, and a negative image $I^n_i$ positioned far away from the anchor image.
\begin{equation}
\label{eq:image_train}
\mathcal{L}_\text{img} = \sum_{I_i^{a,p,n} \in \cal{B}} [d(f(I_i^a), f(I_i^p)) - d(f(I_i^a), f(I_i^n)) + m]_+
\end{equation}
Note that $d(\cdot)$ is the distance function, $f(\cdot)$ is the image branch model, $m$ is the margin, and $[\cdot]_+$ means $ \max \{~0,[\cdot]~\}$. In order to train more efficiently, we find the hardest positive sample with maximal distance and the hardest negative sample with minimal distance to the anchor within the mini-batch $\mathcal{B}$.
\subsection{Cross-modal Local Feature Training}
\label{subsec:local_train}
In this section we describe the second stage of our pipeline, where we pre-train point cloud-based branch using local feature correspondences. The overview can be seen in~\cref{fig:overview}.

\begin{figure*}
  \centering
  \scalebox{0.9}{
  \includegraphics[width=\linewidth]{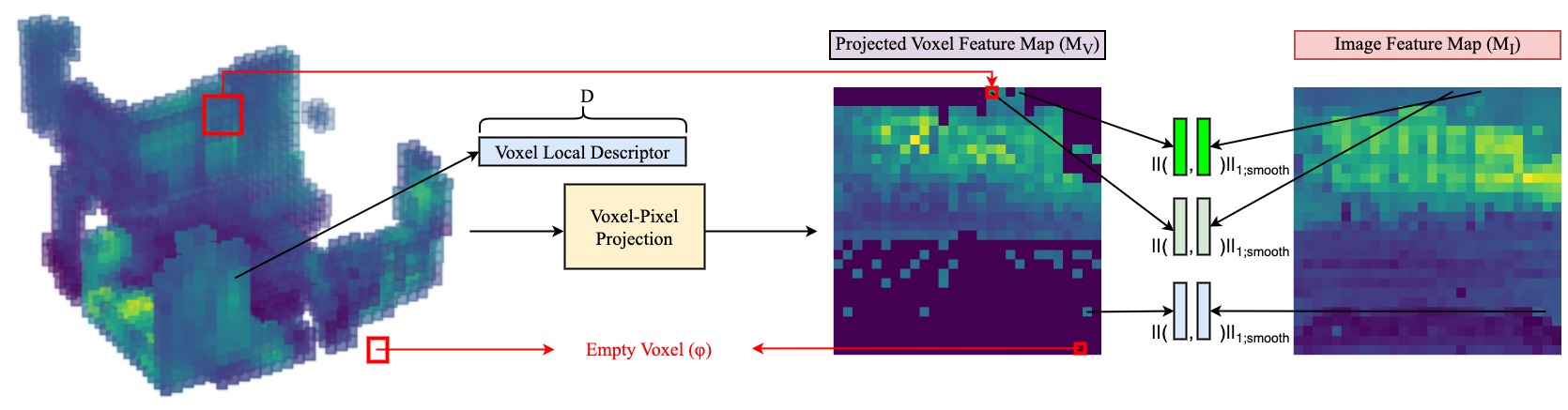}
  }
  \caption{Illustration of our proposed local feature optimization between projected voxel- and image-based feature maps. $\phi$ represents ``empty'' as the 3D feature maps are sparse. Note that the voxel local descriptor is the $\textbf{v}_i^{out}$ introduced in \cref{eq:output_voxel_formulation}. After the projection, multiple $\textbf{v}_i^{out}$ could be projected as per \cref{eq:local_desc_loss}.\vspace{-0.3cm}}
  \label{fig:proj_and_local_descr_opt}
\end{figure*}
\noindent \textbf{Voxel Feature Encoding.} The initial voxel feature $\textbf{v} \in \mathbb{R}^{M \times 3}$ aggregates information from $M$ raw point coordinates contained within the voxel boundaries. We use VoxelNet \cite{zhou2018voxelnet} to extract more detailed descriptor for each voxel $\textbf{v} \in \mathbb{R}^{M \times 3} \quad \rightarrow \quad \mathbb{R}^{D^*}$. Finally, we perform a series of sparse 3D convolutions \cite{yan2018second} to generate a sparse 3D feature map of grid size $28 \times 28 \times 28$, namely $\textbf{V}_{out}$, as formulated in~\cref{eq:output_voxel_formulation}. 
\begin{equation}
    \textbf{V}_\text{out} = \{\textbf{v}_i^\text{out} \in \mathbb{R}^{D}, \textbf{c}_i^\text{out} \in \mathbb{R}^3\}_{1, 2, ..., T^*}
    \label{eq:output_voxel_formulation}
\end{equation}
The $\textbf{v}_i^{out} \in \mathbb{R}^{D}$ represents a local descriptor of a single voxel in the output voxel grid, which is a D-dimensional vector corresponding to the channel size of the 2D feature from $f^{enc}$, while $\textbf{c}_i^{out}$ denotes the coordinate of this voxel within the voxel grid. Here, $\textbf{c}_i^{out}$ is defined with respect to the voxel grid coordinate frame $\{\mathcal{V}\}$. Note that $T^*$ represents the number of non-empty voxel local descriptors. Sparse convolutions allow us to aggregate spatial information from neighboring voxels in a hierarchical fashion, which allows to capture long-distance relations.\\
\noindent \textbf{Voxel-Pixel Projection.} In order to bridge the domain gap between point cloud and image, we introduce simple yet effective {\em Voxel-Pixel Projection} module. This module projects voxels onto the image plane using the pinhole camera model. However, it's important to note that the voxel coordinates are defined within the voxel grid coordinate system denoted as $\{\mathcal{V}\}$. As per~\cref{eq:coor_trans}, we first transform the voxels into the point cloud (LiDAR) coordinate frame and apply projection matrix $\mathbf{M}$ to transform points onto the image plane. This way, we can obtain the voxel-based feature map and establish local descriptor constraints with the image-based features. Projection matrix is assumed to be provided and comprises intrinsic camera parameters and extrinsic LiDAR-camera calibration transformation.
\begin{equation}
\lambda
    \begin{bmatrix}
    u_i \\
    v_i \\
    1
    \end{bmatrix} = 
\mathbf{M} \cdot \left(
\begin{bmatrix}
v_x & 0 & 0 \\
0 & v_y & 0 \\
0 & 0 & v_z
\end{bmatrix}
\textbf{c}_i^{out}+
\begin{bmatrix}
v_x/2 + x_{min} \\
v_y/2 + y_{min} \\
v_z/2 + z_{min}
\end{bmatrix}
\right)
\label{eq:coor_trans}
\end{equation}

\noindent Note that $(v_x, v_y, v_z)$ is the dimension of the output voxel grid and the lower bound of the point cloud range is represented as $(x_\text{min}, y_\text{min}, z_\text{min})$.\\ 

\noindent \textbf{Local Feature Optimization.} During the local descriptor optimization phase shown in \cref{fig:proj_and_local_descr_opt}, we utilize the projected voxel coordinates $(u_i, v_i)$ as indices to retrieve the corresponding local descriptors from the image feature map. Once retrieved, we can apply the local descriptor loss as
\begin{equation}
    \mathcal{L}_\text{local}=\sum_{(u_i,v_i)\in \mathcal{M}_V} || d_i \cdot \mathcal{M}_V(u_i,v_i)  - \mathcal{M}_I(u_i,v_i) ||_\text{1; smooth}.
    \label{eq:local_desc_loss}
\end{equation}
The projected voxel feature map is denoted as $\mathcal{M}_V$, the image feature map is $\mathcal{M}_I$. We also take care of collisions, when multiple voxels are projected to the same pixel, by weighting descriptors with their voxels' inverse depths $d_i$. This way we give preference to the voxels that are closer to the camera, however propagate gradients to all voxels. This strategy allowed more stable training over the z-buffering.

\subsection{Cross-Modal Global Descriptor Training}
\label{subsec:global_train}

In the last stage, we fine-tune the Voxel-Pixel Encoder and train pooling layers together with the subsequent FCN to bring global embeddings closer to their image-based matches with
\begin{equation}
    \mathcal{L}_\text{global}=\sum_i ||f(I_i) - g(P_i)||_\text{1; smooth}
    \label{eq:global_desc_loss},
\end{equation}
where $I_i$ and $P_i$ are an image and a point cloud corresponding to the same location, $f(.)$ and $g(.)$ refer to the corresponding networks. This allows us to ensure global consistency of aggregated descriptors in addition to local similarities enforced in the previous stage.

\section{Experiments and Results}
\label{sec:results}
\subsection{Implementation Details}
\label{sec:implementation_details}
\noindent \textbf{Image Network Training:} 
We resize the image to $224 \times 224$.
During training of the image network, positive pairs are chosen from images that are within 10 meters, while the negative pairs are defined from samples that are more than 25 meters away as \cite{lee20232}. We set the margin of the triplet loss function to 0.3. To handle zero-triplets, i.e. anchor-positive-negative tuples with zero triplet loss, we employ a strategy of gradually increasing the batch size if the proportion of zero-triplets exceeds 30\% of the original batch size. The training with branch expansion rate is adopted from~\cite{komorowski2021minkloc3d} and configured to 1.4, while the maximum batch size is set to 256. 
We use pre-trained model (dino-vits8) and finetune all its parameters together with our GeM + FCN block using \cref{eq:image_train}. A custom batch sampler with at least one positive pair within each batch \{$\mathcal{B}$\} is implemented.
For each sample in \{$\mathcal{B}$\}, its hard positive / negative sample is the farthest / closest sample in \{$\mathcal{B}$\} based on the L2 distance between the global descriptors.\\
\noindent \textbf{Point Cloud Network Training.} We take the fine-tuned image network and freeze all its parameter during the training of the point cloud network. We adopt the following voxelization parameters: point cloud boundaries range is $ x: [0, 44], y: [-22, 22], z: [-4, 18]$, voxel dimensions are set to $[v_x, v_y, v_z] = [0.4, 0.4, 0.2]$. This would allow us to have a final voxel grid with size (110, 110, 110). Both the cost functions in $\mathcal{L}_\text{local}$ and $\mathcal{L}_\text{global}$ are chosen as smooth L1 loss $||\cdot||_\text{1; smooth}$ to ensure robustness to outliers. Adam optimizer and LambdaLR learning rate scheduler are utilized in our training pipeline.
\subsection{Datasets}
\noindent \textbf{Oxford RobotCar Dataset.}
We utilize the Oxford RobotCar  benchmark~\cite{maddern20171} for evaluation, where the same trajectory was traveled over a year in different times of the day and seasonal conditions. We generate data samples following the same protocol as conducted by Cattaneo et al. \cite{cattaneo2020global}, where image is recorded every five meters and the corresponding point cloud map is constructed by concatenating the subsequent 2D LiDAR scans. The four test regions are excluded from the training dataset as per~\cite{uy2018pointnetvlad}.\\
\noindent \textbf{ViViD++ Dataset.} Additionally, we assess the performance of our model on the ViViD++ dataset \cite{lee2022vivid++}, which consists of driving and handheld sequences and offers 3D LiDAR, visual and GPS data. In the scope of our work, we are mainly interested in the urban data, which contains sensor measurements recorded during a day, evening and night. We follow the training procedures proposed by Lee et al. \cite{lee20232} where only the \textit{day1} sequences are used for training, while performing evaluation with \textit{day2} and \textit{night} sequences.\\
\noindent \textbf{KITTI Odometry Dataset.} 
We further test the generalization capability of our VXP  on the KITTI Odometry benchmark \cite{geiger2012we}, which contains sequences with LiDAR scans, images, and ground-truth poses.
\subsection{Results}
\label{sec:results}
Across various datasets we evaluate different combinations of modalities for query and database: 2D-3D (image query and point cloud database), 3D-2D (point cloud query and image database) and their uni-modal variations, i.e. 2D-2D (image-only) and 3D-3D (point cloud-only).\\
\noindent \textbf{Oxford RobotCar.} We adhere to the evaluation metric employed by Cattaneo~\cite{cattaneo2020global}, in which we select each pair of distinct runs from 23 sequences as query and database. The query contains samples only from the four excluded regions as per~\cite{uy2018pointnetvlad}, while database consists of samples from the entire trajectory. Finally, the average of the recall is computed for all the pairs.
In~\cref{tab:recall_oxford_1percent}, we compare our model with the existing cross-modal retrieval approaches, such as the method by Cattaneo et al.~\cite{cattaneo2020global},  $LC^2$ \cite{lee20232} and LIP-Loc \cite{shubodh2024lip}. As the code from Cattaneo et al. \cite{cattaneo2020global} is not publicly released, we have implemented the approach with the authors' help to the best of our abilities. We report performance on different modality configurations, namely database and query combinations of 2D images and 3D point clouds.\\
\noindent Our method outperforms other baselines on 2D-3D place recognition by a significant margin due to the proposed local constraints. We also demonstrate the best performance in the uni-modal retrieval. Fig.~\ref{fig:teaser_fig} shows average recall up to $K = 25$ nearest neighbors for cross-modal place recognition on the Oxford dataset. Our method is the most accurate and precise with respect to all the baselines from Cattaneo et al.~\cite{cattaneo2020global} and LIP-Loc \cite{shubodh2024lip} and across the whole K-range, which validates consistency of our method.
\begin{table}[tb]
  \centering
  \begin{tabular}{@{}l|cccc@{}}
    \toprule
    Recall@1\% & 2D-3D & 3D-2D & 2D-2D & 3D-3D \\
    \midrule
    Cattaneo's~\cite{cattaneo2020global} & 77.3 & 70.4 & 96.6 & 98.4 \\
    $LC^2$~\cite{lee20232} & 81.2 & 73.8 & 84.1 & 83.0\\
    LIP-Loc~\cite{shubodh2024lip} & 77.8 & 73.6 & 90.2 & 92.3\\
    \toprule
    VXP (Ours) & \textbf{84.4} & \textbf{76.9} & \textbf{98.8} & \textbf{98.8} \\
    \bottomrule
  \end{tabular}
\caption{Retrieval performance compared with existing cross-modal methods on Oxford dataset. Our model consistently outperforms other baselines on both cross- and uni-modal settings. }
  \label{tab:recall_oxford_1percent}
\end{table}

\begin{table}
  \centering
  \scalebox{1}{
  \begin{tabular}{l|cc|cl}
    \toprule
      & \multicolumn{2}{c|}{2D-2D} & \multicolumn{2}{c}{3D-3D}\\
      & 1 & 1\% & 1 & 1\% \\
    \midrule
    AnyLoc~\cite{keetha2023anyloc} & 93.5 & 98.9 & -- & -- \\
    MixVPR~\cite{ali2023mixvpr} & 92.8 & 97.7 & -- & -- \\
    MinkLoc3D-S~\cite{zywanowski2021minkloc3dsi} & -- & -- & 95.8 & 99.0 \\
    CASSPR~\cite{Xia_2023_ICCV} & -- & -- & 94.7 & 98.4 \\
    \toprule
    VXP (Ours) & 92.0 & 98.8 & 94.7 & 98.8 \\
    \bottomrule
    \end{tabular}}
  \caption{Retrieval performance compared with existing uni-modal methods on Oxford dataset.  Provided values correspond to Recall@1 and 1\%. Our model has comparable performance with the uni-modal state-of-the-art approaches. \vspace{-0.4cm}}
  \label{tab:oxford_unimodal}
 \end{table}
\noindent We also compare the performance of our method with the state-of-the-art uni-modal approaches for visual place recognition methods AnyLoc~\cite{keetha2023anyloc} and MixVPR~\cite{ali2023mixvpr}, and LiDAR-based retrieval, such as MinkLoc3D-S~\cite{zywanowski2021minkloc3dsi} and CASSPR~\cite{Xia_2023_ICCV}. \cref{tab:oxford_unimodal} shows our method performs on-par with the uni-modal baselines, while additionally offering cross-modal capabilities that are practical for multi-sensor on-board suites.\\ 
\noindent \textbf{ViViD++.} We further evaluate our model on the ViViD++ dataset and compare the results of different approaches on \textit{day1}$-$\textit{day2} sequences in~\cref{tab:vivid_day1_day2}. Note that \textit{day1}$-$\textit{day2} represents query from \textit{day1} sequences and database using \textit{day2} sequences. Overall, we outperform the other baselines \cite{cattaneo2020global, lee20232, shubodh2024lip} on the cross-modal place recognition and perform on par with~\cite{cattaneo2020global} on uni-modal retrieval task.\\
We also evaluate our method on the night-day retrieval, where database map is recorded in the day and queries are obtained at night. We report average performance computed on the \textit{city night}$-$\textit{city day2} and \textit{campus night}$-$\textit{campus day2} sequences from the dataset.
Despite significant appearance differences between night queries and map samples recorded during the day, our VXP is able to tackle this challenge by incorporating information from the LiDAR scans that are not affected by insufficient lighting conditions. As shown in~\cref{tab:vivid_night_day2}, image retrieval (2D-2D) struggles in the challenging scenarios of the night-day retrieval, while cross-modal recognition is capable to offer more accurate place recognition performance across all baselines. Moreover, our approach outperforms other methods such as $LC^2$, ~\cite{cattaneo2020global}, and ~\cite{shubodh2024lip} on the 3D-2D place recognition task and shows highly accurate results based on the top selected retrieval candidate. Specifically, on Recall@1 we achieve a boost in performance by a large margin ($\sim 25 \%$ improvement), which demonstrates the effectiveness of our pipeline for this challenging scenario.

\begin{table}
  \centering
  \scalebox{0.75}{
  \begin{tabular}{l|cc|cc|cc|cl}
    \toprule
     & \multicolumn{2}{c|}{2D-3D} & \multicolumn{2}{c}{3D-2D} & \multicolumn{2}{c|}{2D-2D} & \multicolumn{2}{c}{3D-3D}\\
     & 1 & 1\% & 1 & 1\% & 1 & 1\% & 1 & 1\% \\
    \midrule
    $LC^2$~\cite{lee20232} & 60.9 & 96.0 & 51.8 & 94.6 & 69.2 & 96.9 & 58.1 & 96.1 \\
    Cattaneo's \cite{cattaneo2020global} & 87.6 & 99.6 & 78.6 & 98.6 & 93.4 & 99.8 & 91.0 & \textbf{99.9} \\
    LIP-Loc \cite{shubodh2024lip} & 73.7 & 98.4 & 54.9 & 93.0 & 61.1 & 94.0 & 78.8 & 97.4 \\
    \toprule
    VXP (Ours) & \textbf{96.8} & \textbf{99.6} & \textbf{94.7} & \textbf{99.8} & \textbf{96.7} & \textbf{99.9} & \textbf{97.0} & 99.7 \\
    \bottomrule
    \end{tabular}}
  \caption{Retrieval performance (average recall) for top 1 and 1\% retrieved places on the ViViD++ dataset (\textit{day1}$-$\textit{day2} sequences). Our model outperforms the other baselines on both uni- and cross-modal experiments.}
  \label{tab:vivid_day1_day2}
 \end{table}
 \begin{table}
  \centering
  \begin{tabular}{@{}l|cc|cl@{}}
    \toprule
    & \multicolumn{2}{c|}{2D-2D}  & \multicolumn{2}{c}{3D-2D} \\
    & 1 & 1\% & 1 & 1\% \\
    \midrule
    $LC^2$ \cite{lee20232}& 0.8 & 5.5 & 49.4 & 93.4\\
    Cattaneo's \cite{cattaneo2020global} & 2.2 & 10.1 & 56.9 & 94.9 \\
    LIP-Loc \cite{shubodh2024lip} & 2.7 & 12.0 & 45.5 & 90.0\\
    \toprule
    VXP (Ours) & \textbf{10.2} & \textbf{21.7} & \textbf{82.0} & \textbf{97.5}\\
    \bottomrule
  \end{tabular}
  \caption{Retrieval performance (average recall) for top 1 and 1\% retrieved places on ViViD++ dataset (\textit{night}$-$\textit{day2} sequences). We can observe the advantage of deploying LiDAR scans as a query, which significantly boosts performance for all baselines. Due to the proposed architectural design, our VXP performs the best in both settings.\vspace{-0.3cm}}
  \label{tab:vivid_night_day2}
\end{table}
\noindent \textbf{KITTI Odometry Benchmark.} The results are shown in \cref{tab:kitti}. Different to the evaluation procedure followed by \cite{Xia_2023_ICCV, cattaneo2022lcdnet} for LiDAR-based place recognition, we propose our own evaluation protocol on the dataset. Specifically, we train the model on sequences 03, 04, 05, 06, 07, 08, 09, 10. For testing we select 4 regions from sequences 00 and 02 and include the remaining parts of the trajectory into the training data. Notably, none of the sequences traverses the same place, so we test our model on completely unseen regions to demonstrate generalisation capability of our method. Further training details are provided in the supplementary.\\
\noindent As shown in~\cref{tab:kitti} our method demonstrates competitive performance on all configurations. Since the full code for the $LC^2$ was not publicly available at the submission time, we could not provide comparison on this benchmark. While LIP-Loc~\cite{shubodh2024lip} achieves the best performance on Recall@1\% 2D-3D setting, it is more sensitive to the sampling range of the database samples and queries. We provide details of the experiment in the supplementary.

\begin{table}
  \centering
  \scalebox{0.75}{
  \begin{tabular}{@{}l|cc|cc|cc|cc@{}}
    \toprule
     & \multicolumn{2}{c}{2D-3D} & \multicolumn{2}{|c}{3D-2D} & \multicolumn{2}{c|}{2D-2D} & \multicolumn{2}{c}{3D-3D} \\
     & 1 & 1\% & 1 & 1\% & 1 & 1\% & 1 & 1\%\\ 
    \midrule
    Cattaneo's \cite{cattaneo2020global} & 15.9 & 23.4 & 12.8 & 28.7 & 95.7 & 97.8 & 58.6 & 71.3\\
    LIP-Loc \cite{shubodh2024lip} & 20.0 & \textbf{40.9} & \textbf{21.9} & 29.3 & 29.3 & 44.0 & 27.2 & 37.8\\
    \toprule
    VXP (Ours) & \textbf{32.1} & 38.6 & \textbf{36.1} & \textbf{38.3} & \textbf{97.8} & \textbf{100.0} & \textbf{86.3} & \textbf{89.4} \\
    \bottomrule
  \end{tabular}}
  \caption{Retrieval performance (average recall) for top 1 and 1\% retrieved places on KITTI Odometry dataset (00, 02 sequences). Our model shows competitive performance among all baselines.  }
  \label{tab:kitti}
\end{table}

\begin{table}
    \centering
    \begin{tabular}{@{}l|cccc@{}}
    \toprule
      & \multicolumn{2}{c}{2D-3D} & \multicolumn{2}{c}{3D-2D} \\
    & 1 & 1\% & 1 & 1\%\\
    \midrule
    Global-only & 41.3 & 81.5 & 30.2 & 74.7 \\
    \toprule
    Local + Global (Ours) & \textbf{44.6} & \textbf{84.4} & \textbf{30.8} & \textbf{76.9}\\
    \bottomrule
   \end{tabular}
   \caption{Ablation study of the local feature optimization (\cref{eq:local_desc_loss}) for cross-modal retrieval on the Oxford RobotCar benchmark. Introducing local constraints significantly improves retrieval accuracy over global-only baseline (\cref{eq:global_desc_loss}), which validates our architectural design. \vspace{-0.3cm}}
   \label{tab:w_wo_two_stage_opt}
\end{table}

\section{Ablation Studies}
\textbf{Local Descriptor Loss Analysis.}
\label{subsec:multistage}
We evaluate the impact of the Local Descriptor Optimization (\cref{subsec:local_train}) on the cross-modal place recognition. As shown in \cref{tab:w_wo_two_stage_opt}, the proposed combination of local and global optimizations allows the model to effectively bridge the domain gap between image and point cloud and achieve higher cross-modal retrieval performance.

\begin{figure}[t]
  \centering
   \includegraphics[width=1\linewidth]{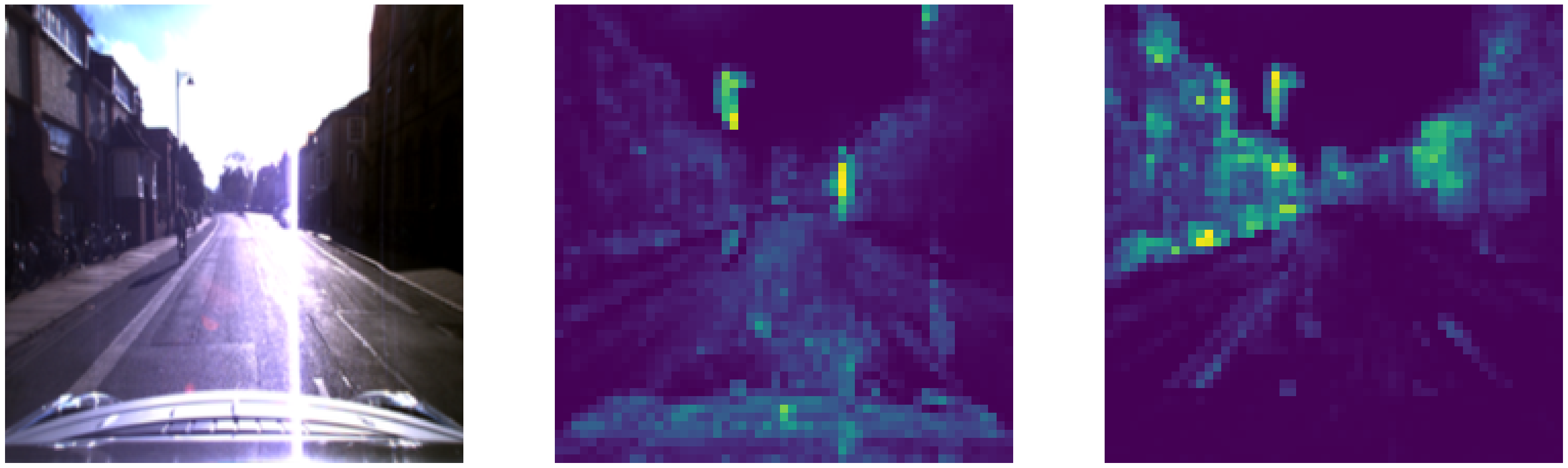}
   \caption{DINO fine-tuning effects on attention maps. From left to right: an input image, an attention map generated by pretrained DINO's ViTs-8 without fine-tuning and a map produced after fine-tuning. Due to the latter, important scene structures such as buildings and traffic poles receive higher attention.}
   \label{fig:att_com}
\end{figure}

\noindent \textbf{Fine-tuning Image Backbone.}
Foundation models such as DINO~\cite{oquab2023dinov2} have demonstrated capability of addressing a wide range of tasks~\cite{amir2021deep}. However, we have noticed that their off-the-shelf performance on the visual (2D-2D) place recognition task is quite poor and fine-tuning is necessary to reach a competitive accuracy. Specifically, we scored only 59.5\% on the 2D-2D Recall@1 with the pre-trained DINO ViTs-8 model, while with additional fine-tuning we achieved 2D-2D accuracy of 92.0\% (\cref{tab:oxford_unimodal}). We can also observe the effect of fine-tuning the model on the attention maps. An example from the Oxford benchmark is shown in \cref{fig:att_com}. Specifically, buildings, road markings and traffic lights receive higher attention scores after fine-tuning, while the car hood is ignored.

\noindent \textbf{Voxel-Pixel Projection Module Analysis.} 
We compare our VXP model against a simple baseline, Ortho-VXP, which transforms $\textbf{V}_\text{out}$ to a dense form and performs an orthographic projection of the features to obtain an analogue in the image plane.
As shown in \cref{tab:oxford_wo_proj}, we achieve a boost on the cross-modal localization due to the perspective nature of the VXP module, which associates voxels to corresponding pixels considering the former depth and provides stronger place recognition cues than maintaining original distances and size as per orthographic projection.

\begin{figure}
  \centering
  \includegraphics[width=0.95\linewidth]{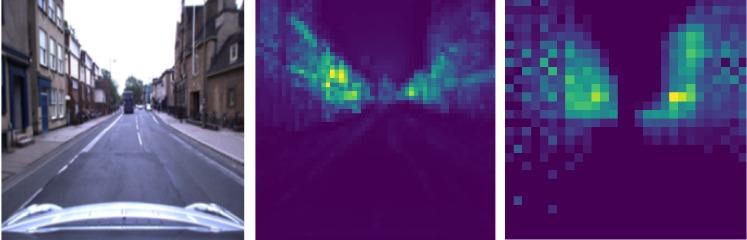}
  \caption{From left to right: an input image, its attention map and projected feature map generated from the respective point cloud.\vspace{-0.3cm}}
  \label{fig:com_rgb_att_invd}
\end{figure}

\begin{table}
  \centering
  \begin{tabular}{@{}l|cccc@{}}
    \toprule
     Recall@1\% & 2D-3D & 3D-2D & 2D-2D & 3D-3D \\
    \midrule
    Ortho-VXP & 78.1 & 71.9 & \textbf{98.8} & \textbf{98.9}\\
    \toprule
    VXP (Ours) & \textbf{84.4} & \textbf{76.9} & \textbf{98.8} & 98.8\\
    \bottomrule
  \end{tabular}
  \caption{Ablation study of projection module on the Oxford RobotCar dataset. Perspective projection with VXP benefits localization when compared with its orthographic analog, Ortho-VXP. \vspace{-0.4cm}
  }
  \label{tab:oxford_wo_proj}
\end{table}
\noindent \textbf{Qualitative Evaluation for VXP.}
As we have shown in \cref{sec:results}, VXP achieves state-of-the-art cross-modal retrieval performance and maintains high uni-modal global localization accuracy. At the same time, we are capable of mitigating the domain gap between different modalities and learning expressive shared latent space. We demonstrate the correlation between the attention map of an RGB image and the feature map of the projected voxels in \cref{fig:com_rgb_att_invd}. Notably, the projected voxels exhibit a similar pattern with the image-based attention map. Since our focus is on place recognition, structures such as buildings carry greater significance, resulting in higher attention scores in those regions for feature maps from both modalities. With this, global descriptors are learned based on consistent information across modalities and we are capable of effectively bridging the domain gap.

\noindent \textbf{Training and Inference Efficiency.}
We evaluate model inference time using a single RTX3080 and pre-processing time with Intel i7-12700. Depth image generation for $LC^2$ \cite{lee20232} baseline is done on GPU. Our VXP takes 7 ms to obtain a global descriptor for an image and 18 ms for a point cloud respectively, while $LC^2$ \cite{lee20232} encodes input image in 17 ms and LiDAR scan in 53 ms due to expensive pre-processing step of depth image generation and point cloud-to-range image conversion. In terms of model parameters and memory footprint, 2D and 3D networks of VXP have 21.7M (87.2MB) and 5.9M (23.6MB) parameters respectively. With this, our model is fast and lightweight to run as part of a real-time system. Notably, the reference map can be encoded offline.

\section{Limitations and Future Work}
Our VXP pipeline comprises three steps as described in~\cref{sec:vxp}. Although this multi-stage design showcases the best performance based on our ablation studies (\cref{subsec:multistage}), end-to-end training requires less engineering effort and opens a possibility for generalization when training on larger or multi-source datasets, which is desirable for the autonomous driving applications. In addition, our model is specific for every dataset. While it achieves good performance on the unseen views from the training in-domain dataset, it does not work on different, out-domain sequences. As VXP needs a dataset-specific calibration matrix to establish local descriptor consistency, it remains a limitation towards multi-dataset generalization. Learning calibration on diverse input images and point clouds is a straightforward extension of the VXP pipeline, which is part of the future work. 
\section{Conclusion}
We have presented a new framework, Voxel-Cross-Pixel (VXP), for camera-LiDAR place recognition. VXP makes use of a novel 3D-to-2D projection module specifically designed to establish local feature correspondences and facilitate bridging the domain gap between LiDAR scans and images. To this end, we proposed a cross-modal pipeline, which captures both fine-grained local details and broader global context. Notably, our approach directly works on raw data without any pre-processing steps. Experimental evaluations demonstrate that VXP provides a new state-of-the-art performance on cross-modal image-LiDAR retrieval and offers competitive performance against uni-modal baselines. It shows real-time capability and low memory footprint, which makes it an excellent candidate for deployment on the embedded systems.

\clearpage
\setcounter{page}{1}
\maketitlesupplementary
In the supplementary we provide details on the used coordinate system convention in~\cref{sec:coord}, evaluation procedure on the KITTI Odometry benchmark (~\cref{sec:eval_kitti}~), and further qualitative results in~\cref{sec:qualit_results}. Moreover, we visualize cross-modal local correspondences in the latent space in~\cref{sec:matches} and report few failure cases in~\cref{sec:fail_cases}. 
\begin{figure}[ht]
  \centering
  \includegraphics[width=1\linewidth]{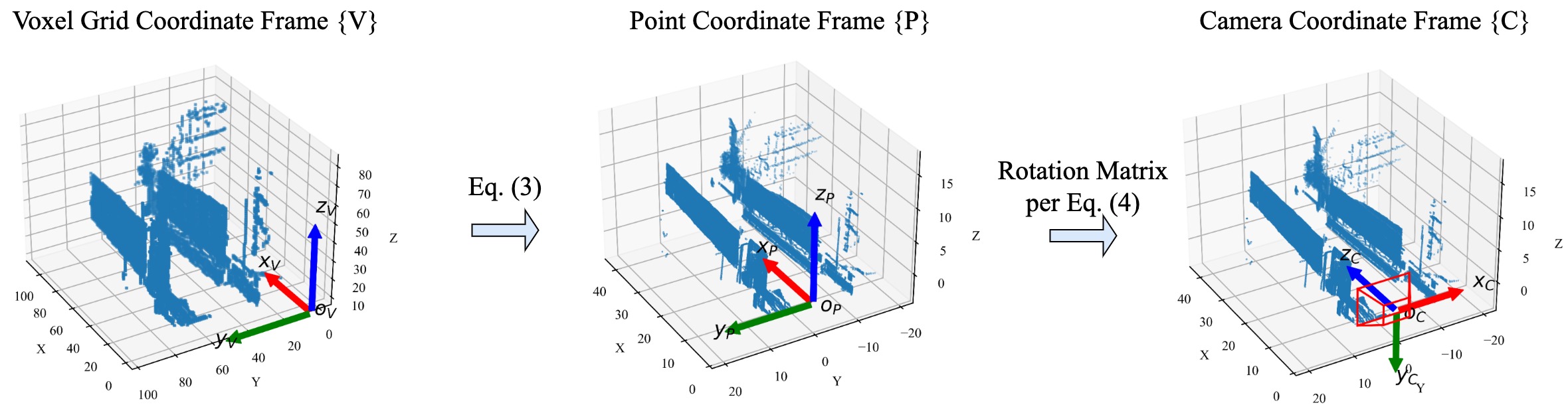}
  \caption{Illustrations of voxel grid coordinate frame $\mathcal{\{V\}}$ (left), intermediate point cloud (LiDAR) system $\mathcal{\{P\}}$ (middle), and target camera coordinate frame $\mathcal{\{C\}}$ (right). Note that once the points are transformed to $\mathcal{\{C\}}$, we can apply pinhole camera projection to project points to pixel coordinate frame.}
  \label{fig:coordinate_frames}
\end{figure}
\section{Coordinate Frames}
\label{sec:coord}
In \cref{subsec:local_train} (main), we introduce Voxel-Pixel Projection module along with the associated coordinate transformations. Accordingly, in \cref{fig:coordinate_frames}, we provide a comprehensive illustration of the operational coordinate frames  $\mathcal{\{V\}}$, $\mathcal{\{P\}}$, and $\mathcal{\{C\}}$ and demonstrate the transformations between them as per \cref{eq:coor_trans} (main). Please note that camera parameters and relative transformation between sensors (camera and LiDAR) are known and provided as part of the datasets (Oxford, ViViD++, and KITTI). When performing pinhole camera projection as per \cref{eq:coor_trans}  in the main paper, normalized intrinsics are used for adapting to different sizes of a feature map.
\section{Training / Testing Setup (KITTI)}
\label{sec:eval_kitti}
\begin{figure}[h!]
  \centering
  \begin{subfigure}{0.48\linewidth}
    \includegraphics[width=1.0\linewidth]{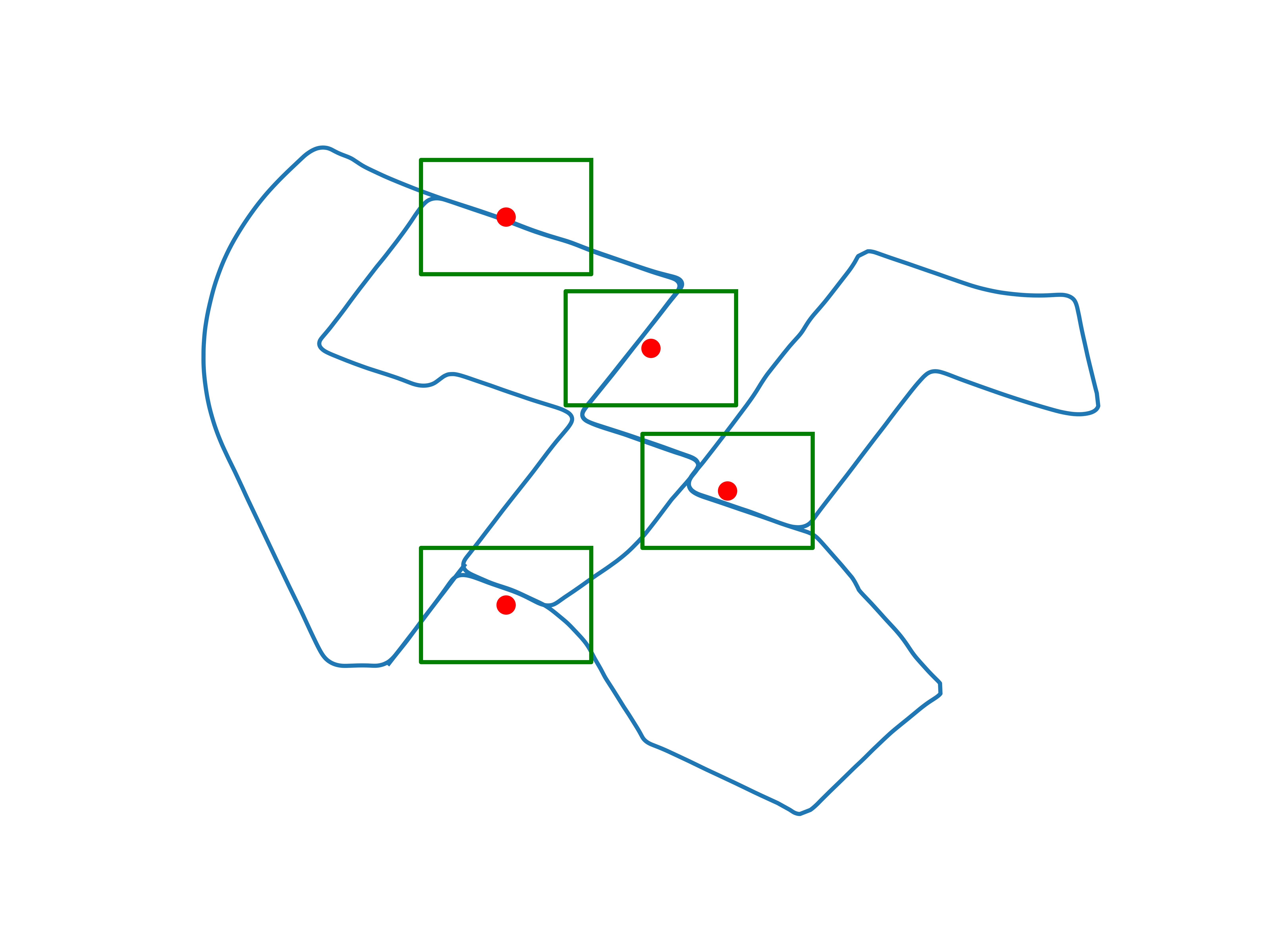}
    \caption{Test regions in KITTI sequence 00.}
    \label{fig:kitti_00}
  \end{subfigure}
  \hfill
  \begin{subfigure}{0.48\linewidth}
    \includegraphics[width=1.0\linewidth]{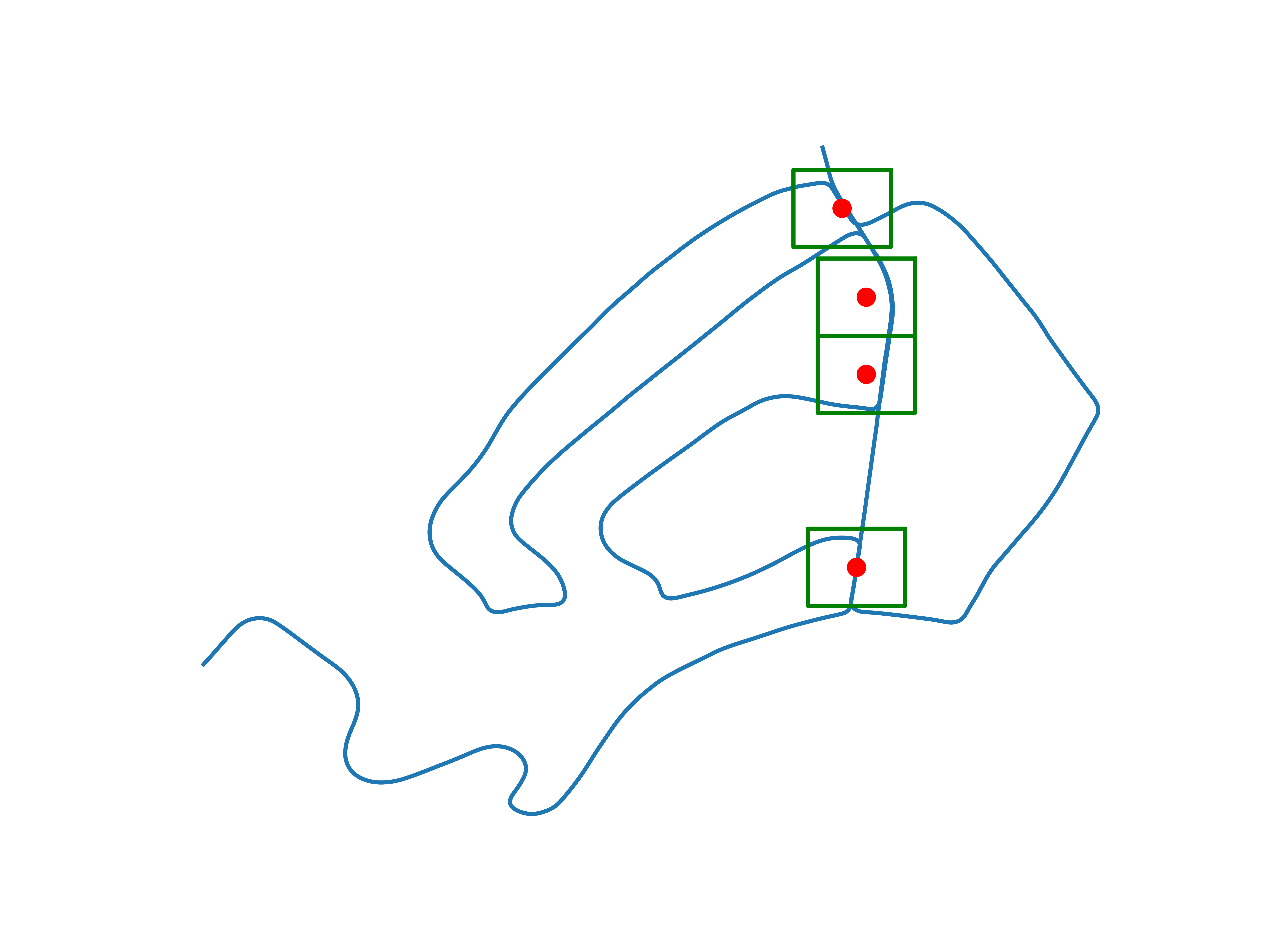}
    \caption{Test regions in KITTI sequence 02.}
    \label{fig:kitti_02}
  \end{subfigure}
  \caption{We select 4 non-overlapping regions from KITTI (sequences 00, 02) and exclude their samples during training. Ability to perform accurate place retrieval in these regions is important to tackle localization drift of SLAM systems.}
  \label{fig:kitti_test_regions}
\end{figure}
In \cref{tab:kitti} (main) we present the evaluation on the KITTI Odometry Benchmark~\cite{geiger2012we}. Our evaluation protocol draws inspiration from the methodology employed in the Oxford RobotCar dataset~\cite{maddern20171}. As shown in \cref{fig:kitti_test_regions}, we define non-overlapping regions in KITTI sequences 00 and 02 and exclude samples from these regions during the training process. This ensures that the model is tested on unseen scenes. Throughout the training process, samples from the test regions in sequence 02 are utilized for validation, and we select the best model based on the validation set. 
During testing, both queries and database are sampled every 20 meters with the start offset of 5 meters. This ensures that samples from queries and database are not repeated, and the positive samples from the constructed query-database pair are in the range of [4.3m, 17.0m] to their corresponding queries as per \cref{tab:kitti_sampling_interval_range}. Notably, to mimic a real-world scenario of detecting loop candidates, we consider a ``revisit'' location by only keeping the positive samples ($D_{t_i}$) of a query ($Q_{t_0}$) such that $t_i < t_0$ and $t_0 - t_i > 10$. In other words, positive samples older than 10 seconds from the query timestamp are all ``revisit'' places. The 10-second threshold is determined empirically based on the sequences 00 and 02.
\begin{table}
  \centering
  \begin{tabular}{@{}lccc@{}}
   \toprule
    & Min. & Max. & Avg. \# Positive\\
    \toprule
    5m & 0.3 & 19.9 & 5.5 \\ 
    10m & 0.6 & 20.0 & 3.2\\ 
    15m & 0.4 & 19.9 & 2.5\\ 
    20m & 4.3 & 17.0 & 1.7\\ 
    \bottomrule
  \end{tabular}
  \caption{Distance range [Min., Max.] of positive samples and their average number per query w.r.t different sampling intervals.} 
  \label{tab:kitti_sampling_interval_range}
\end{table}
\subsection{Sensitivity to Different Sampling (KITTI)}
\begin{figure}
    \includegraphics[width=1\linewidth]{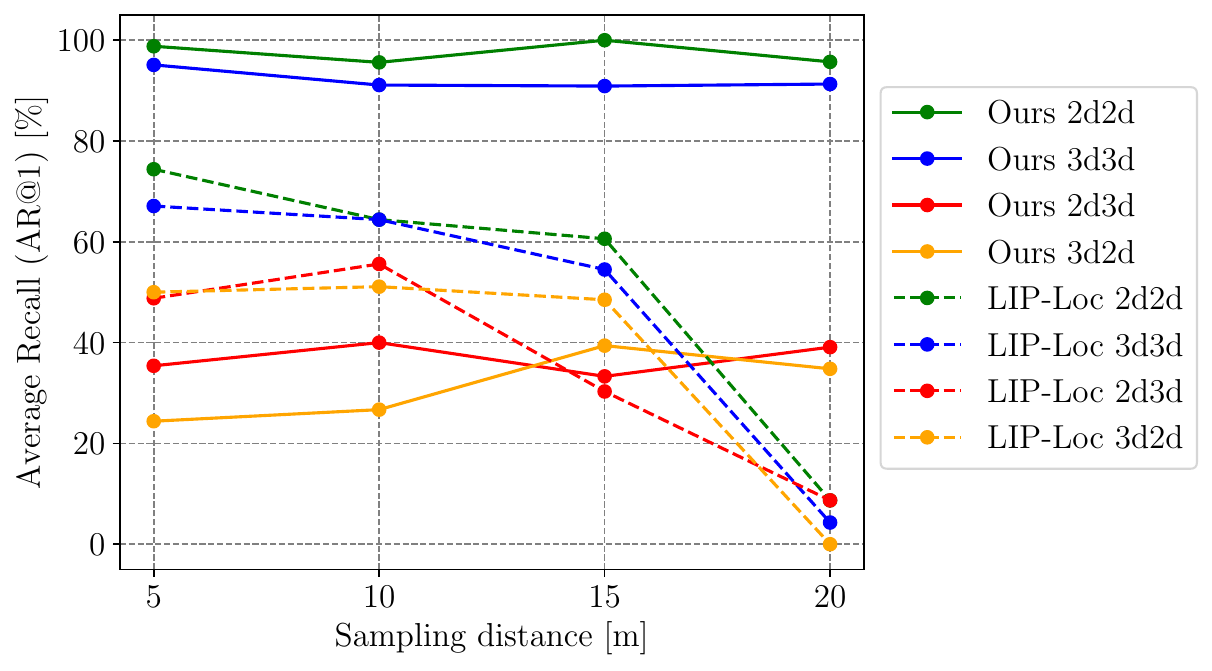}
    \caption{The impact of database-query sampling on VXP and LIP-Loc~\cite{shubodh2024lip} retrieval accuracy. Our VXP shows consistent performance for all sampling ranges, while LIP-Loc results deteriorate rapidly. }
    \label{fig:sensitivitysampling}
\end{figure}

In \cref{sec:results} (main), we discussed the sensitivity of Lip-Loc \cite{shubodh2024lip} retrieval performance to variations in the sampling interval of queries and the database samples.
To further investigate this observation, we perform additional studies in \cref{fig:sensitivitysampling}. As can be seen, our approach remains consistent across different sampling intervals. Lip-Loc, however, exhibits significant performance fluctuations when the sampling interval is changed. 
Notably, reducing the sampling interval results in more samples being classified as positives for each query, with these positive samples being spatially much closer to the query itself, as shown in \cref{tab:kitti_sampling_interval_range}.
This sensitivity in Lip-Loc could be attributed to its training methodology, which uses N-pair batched contrastive loss. In their approach, a pair of an image ($I_{t_i}$) and a LiDAR-scan ($P_{t_j}$) is considered a positive match only when $i=j$, while all the others (when $i \neq j$) are counted as negatives. Therefore, even a LiDAR-scan located 5 meters away from the image would still be labeled as negative, which does not allow the network to generalize to different sensor frequencies and setups. This limitation can be observed from the LIP-Loc performance on the Oxford RobotCar and ViViD++ benchmarks (\cref{tab:recall_oxford_1percent} and \cref{tab:vivid_day1_day2} from main paper, respectively), where camera and LiDAR timestamps are not synchronized, negatively affecting the method's retrieval accuracy.\\
\noindent In contrast, VXP employs a training strategy that learns a shared embedding space by mimicking the output from an image network trained with a triplet loss function. In our approach, samples within 10 meters are considered positive, while those beyond 25 meters are labeled as negative as discussed in \cref{sec:implementation_details} (main). This training strategy enables VXP to robustly retrieve similar locations from the query, even under the more challenging conditions of a 20-meter query-database sampling interval.

\section{Ablation Study of Image Network}
We conducted extensive testing with various image encoders and pooling layers, and the results are summarized in \cref{tab:teacher_ablation}. Overall, it is evident that DINO \cite{caron2021emerging} stands out as the most favorable choice for the image encoder. Concerning the pooling layers, GeM \cite{radenovic2018fine} seems to perform slightly better than NetVLAD \cite{arandjelovic2016netvlad}. Based on the experiments, it is clear that the combination of DINO + GeM + FCN yields the most optimal results.
\begin{table}[tb]
  \centering
  \begin{tabular}{@{}lcc@{}}
    \toprule
  2D-2D &  Recall@1 & Recall@1\% \\ 
    \midrule
V16+N+L   & 80.4   & 91.6\\
V16+G+L   & 81.7   & 92.7\\
R18+N+L   & 77.2   & 90.3\\
R18+G+L   & 78.7   & 91.1\\
Dino+N+L   & \textbf{85.3}   & 94.9\\
\toprule
Dino+G+L (Ours)   & \textbf{85.3}   & \textbf{95.0} \\
    \bottomrule
  \end{tabular}
  \caption{The comparison of the different combinations of image encoder and pooling layer. V16 represents VGG16 \cite{simonyan2014very}, R18 is ResNet18 \cite{he2016deep}, Dino is DINO's ViTs-8 \cite{caron2021emerging}, N is NetVLAD \cite{arandjelovic2016netvlad}, G refers to GeM \cite{radenovic2018fine}, L means fully-connected layer. Our architectural design yeilds the best 2D-2D performance.}
  \label{tab:teacher_ablation}
\end{table}
\begin{figure*}[ht]
  \centering
  \begin{subfigure}{0.5\linewidth}
    \includegraphics[width=1\linewidth]{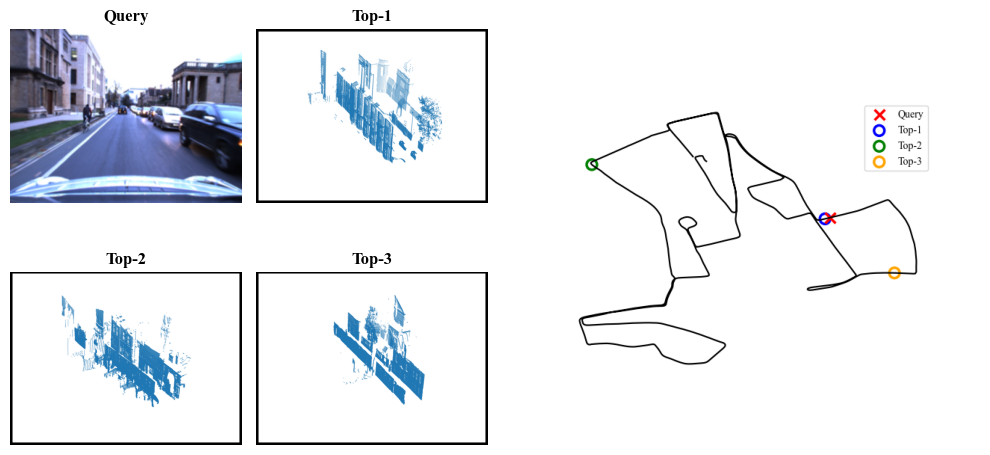}
    \caption{2D-3D retrieval}
    \label{fig:oxford2d3d}
  \end{subfigure}%
  \begin{subfigure}{0.5\linewidth}
    \includegraphics[width=1\linewidth]{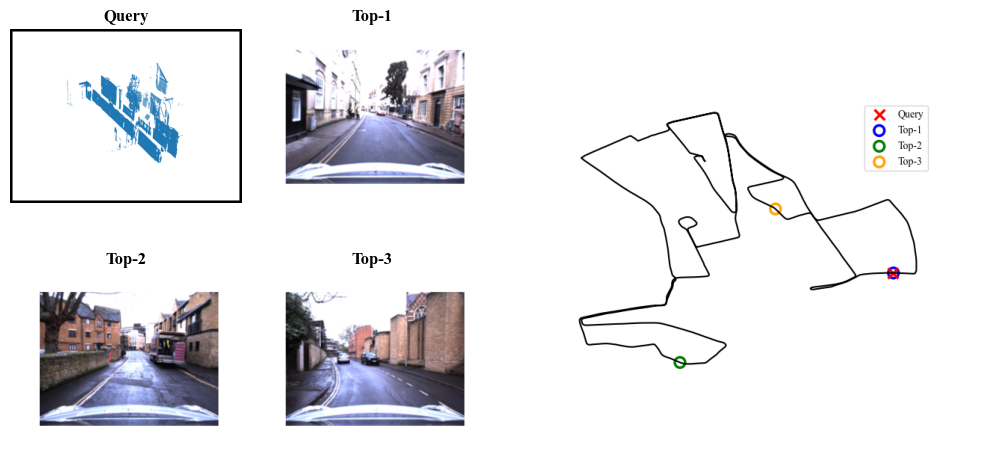}
    \caption{3D-2D retrieval}
    \label{fig:oxford3d2d}
  \end{subfigure}
  \caption{Qualitative results for 2D-3D and 3D-2D cross-modal retrieval of our VXP on the Oxford RobotCar benchmark. We demonstrate the query and the top 3 closest places retrieved from a given map by our method. While the database samples traverse the same route, they are captured at different times when the environmental conditions vary. We can observe that the top 1 candidate is spatially close to the query, demonstrating our approach's effectiveness and accuracy.}
  \label{fig:oxford_traj_recall}
\end{figure*}
\section{Qualitative Results in Challenging Illumination Conditions}
\label{sec:qualit_results}
\begin{figure*}
  \centering
  \begin{subfigure}{0.5\linewidth}
    \includegraphics[width=1\linewidth]{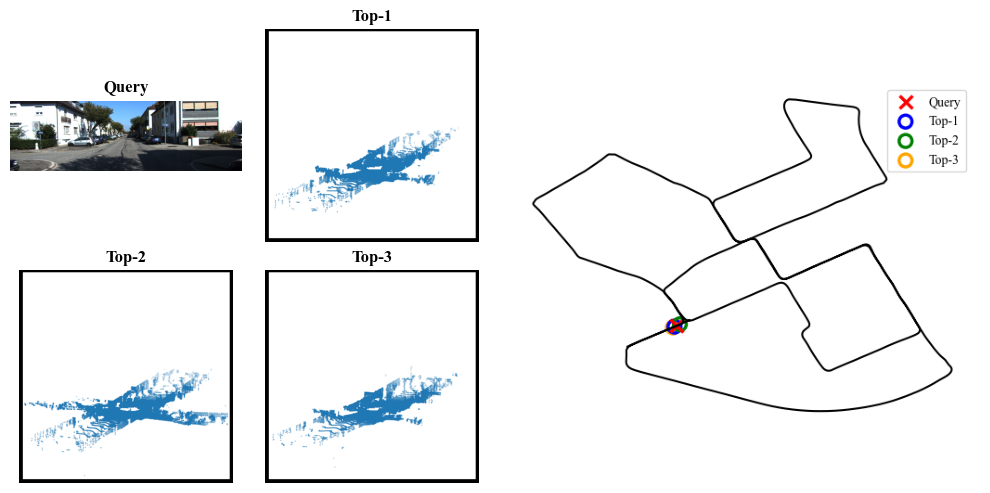}
    \caption{KITTI (00) 2D-3D}
  \vspace{-5pt}
    
    \label{fig:kitti2d3d}
  \end{subfigure}%
  \begin{subfigure}{0.5\linewidth}
    \includegraphics[width=1\linewidth]{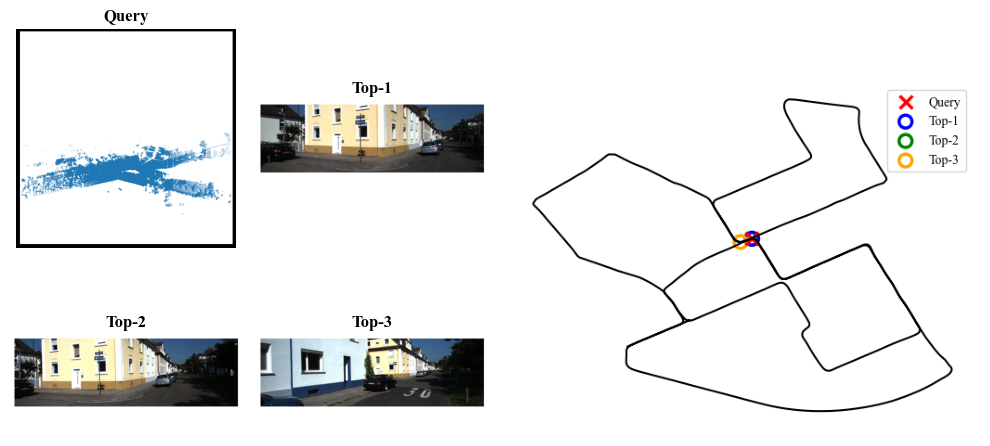}
    \caption{KITTI (00) 3D-2D}
  \vspace{-5pt}
    \label{fig:oxford3d2d}
  \end{subfigure}
  \caption{Qualitative results for 2D-3D and 3D-2D cross-modal retrieval of our VXP on the KITTI Odometry benchmark. We demonstrate the query and the top 3 closest places retrieved from a given map by our method. As described in \cref{sec:eval_kitti} test queries are taken from the region unseen during training. We can observe that all top 3 candidates are spatially close to the query, demonstrating our approach's effectiveness and accuracy.}
  \label{fig:kitti_traj_recall}
  \vspace{-15pt}
\end{figure*}

\begin{figure*}[t]
  \centering
  \begin{subfigure}{0.5\linewidth}
    \includegraphics[width=1\linewidth]{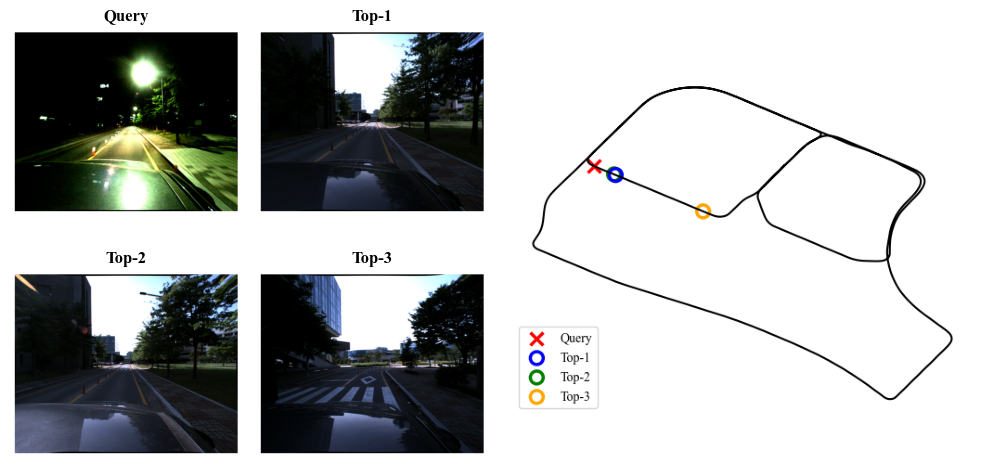}
    \caption{ViViD++ campus night-day2 2D-2D failed.}
    \label{fig:vividcampus2d2dfail}
  \end{subfigure}%
  \begin{subfigure}{0.5\linewidth}
    \includegraphics[width=1\linewidth]{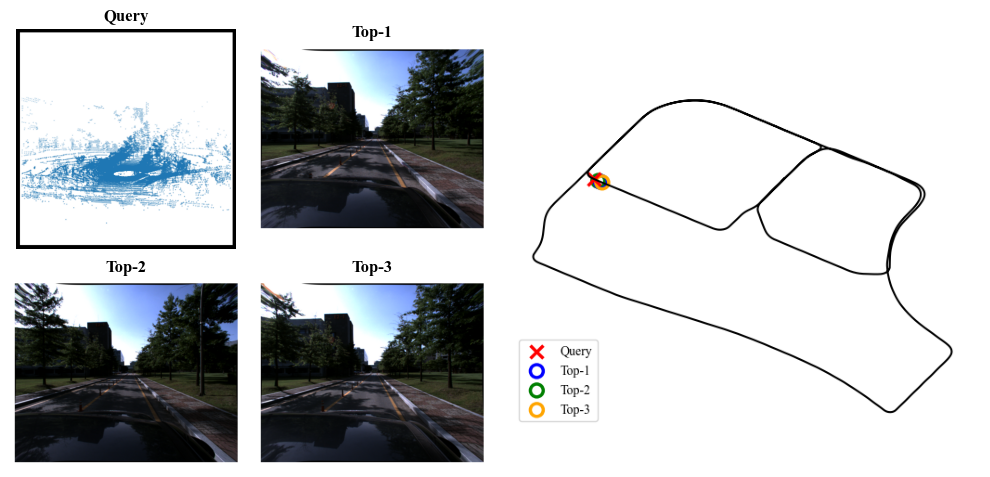}
    \caption{ViViD++ campus night-day2 3D-2D succeeded.}
    \label{fig:vividcampus3d2dsuccess}
  \end{subfigure}
  \begin{subfigure}{0.5\linewidth}
    \includegraphics[width=1\linewidth]{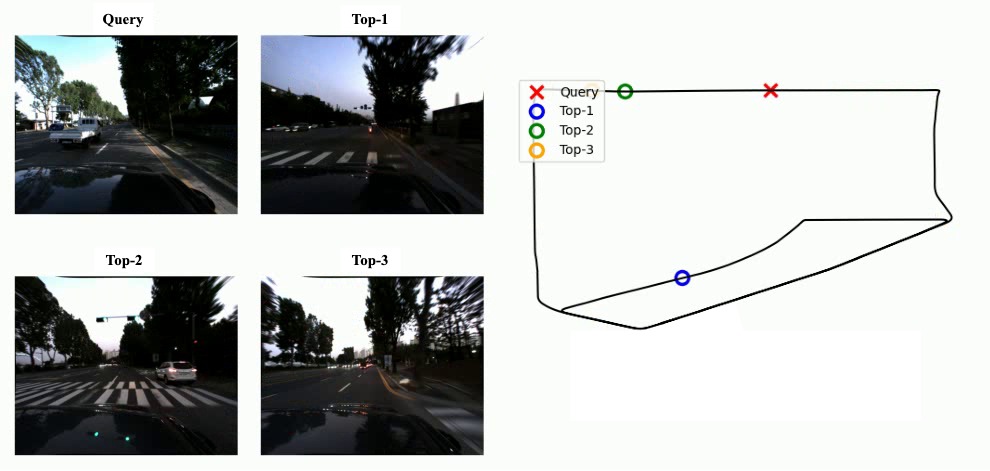}
    \caption{ViViD++ city day1-evening 2D-2D failed.}
    \label{fig:vividcity2d2dfail}
  \end{subfigure}%
  \begin{subfigure}{0.5\linewidth}
    \includegraphics[width=1\linewidth]{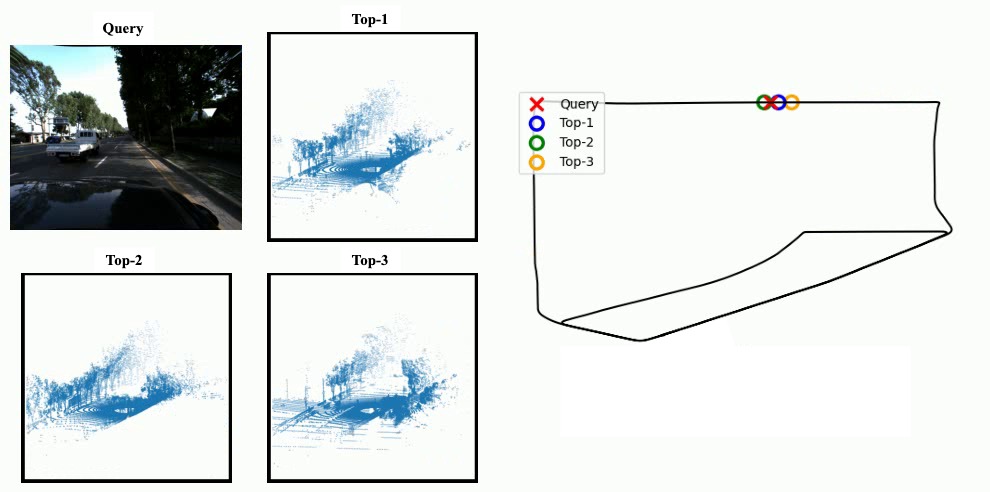}
    \caption{ViViD++ city day1-evening 2D-3D succeeded.}
    \label{fig:vividcity2d3dsuccess}
  \end{subfigure}
  \caption{Qualitative results of uni-modal (left) and cross-modal (right) on challenging illumination conditions such as evening and night sequences from the ViViD++ dataset. While 2D-2D place recognition is impaired by poor illumination conditions, integration of LiDAR scans, which remain unaffected, mitigates the issue and allows accurate cross-modal retrieval. }
  \label{fig:vivid_traj_recall}
\end{figure*}

Given that cameras are sensitive to changes in illumination conditions, robust visual place recognition at night poses significant challenges. In this experiment, we qualitatively demonstrate the differences in image-only retrieval (2D-2D) against cross-modal (2D-3D and 3D-2D) place recognition for the task of day-night and day-evening place recognition with reduced visibility.\\
\noindent In~\cref{fig:vividcity2d2dfail} we can observe that the top 3 places for 2D-2D retrieval are quite far from the query. This suggests that image-based place recognition struggles to retrieve the correct candidate when the images are under different illumination conditions.
However, VXP successfully retrieves the closest candidates using 2D-3D cross-modal retrieval as illustrated in \cref{fig:vividcity2d3dsuccess}. 
This capability could explain why VXP exhibits slightly better top-1 2D-3D recall performance than its 2D-2D counterpart in \cref{tab:vivid_day1_day2} (main). 
Notably, VXP stands out as the only model (compared to \cite{cattaneo2020global, lee20232}) emphasizing the practical advantage of cross-modal retrieval with respect to the uni-modal counterpart.\\
\noindent We demonstrate additional qualitative results of the cross-modal retrieval task on Oxford RobotCar and KITTI datasets in \cref{fig:oxford_traj_recall} and \cref{fig:kitti_traj_recall} respectively. These observations underscore the versatility and effectiveness of cross-modal retrieval approaches in challenging real-world scenarios.
\begin{figure}[ht]
  \centering
  \begin{overpic}[width=1\linewidth,tics=5,]
  {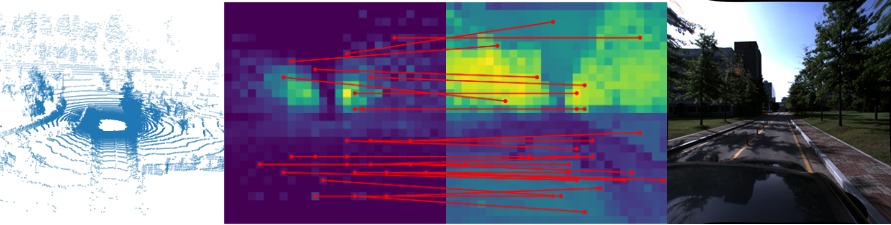}
  \put(5,-2.5){{\tiny LiDAR point cloud }}
  \put(26,-2.5){{\tiny Projected voxel feature map }}
  \put(54,-2.5){{\tiny Image feature map }}
  \put(78,-2.5){{\tiny Corresponding image }}
  \end{overpic}
  \caption{Example of local feature correspondences (red) between a projected voxel feature map and an image feature map on the ViViD++ dataset. Cross-matches are established by minimizing the cosine similarity distance between learned descriptors. Being important landmarks for the place recognition task, buildings and trees receive fairly accurate local matches. }
  \label{fig:correspondences}
\end{figure}
\section{Correspondences in Local Feature Space}
\label{sec:matches}
We visualize cross-modal matches from the ViViD++ dataset~\cite{lee2022vivid++} in \cref{fig:correspondences} established by picking the closest pairs of learned local descriptors. 
It is worth noting that LiDAR scans and images do not capture exactly the same information about the scene since LiDAR and camera are not synchronized. In addition, while we utilize data recorded by traversing the same route, the distance between samples corresponding to the same location can be significant among different traversals. For instance, in the ViViD++ dataset, the distance between an image and the corresponding point cloud averages about 7 meters according to calibration and GPS/INS poses.\\
As it can be seen from \cref{fig:correspondences}, feature correspondences stemming from trees and building structures are accurately established. However, we encounter challenges in regions where voxel features are projected onto the ground region due to the ambiguous nature of respective image features. Therefore, capturing reliable correspondences within the ground regions appears to be a difficult task. Since the ground is constantly present in driving sequences, this misalignment only has minimal impact on the distinctiveness of the estimated global descriptors. Instead, it leverages correctly aligned correspondences from buildings and other static distinct objects in the scene to achieve state-of-the-art cross-modal performance.

\section{Failure Cases}
\begin{figure*}
  \centering
  \begin{subfigure}{0.5\linewidth}
    \includegraphics[width=1\linewidth]{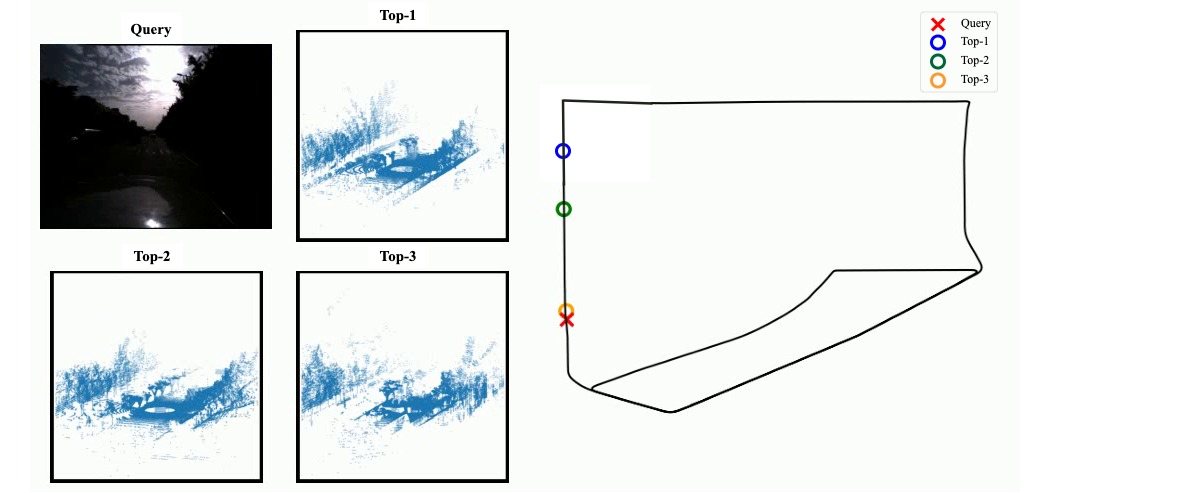}
    \caption{ViViD++ City day1-day2 2D-3D failed.}
  \vspace{-5pt}
    
    \label{fig:vividcity2d3dfail}
  \end{subfigure}%
  \begin{subfigure}{0.5\linewidth}
    \includegraphics[width=1\linewidth]{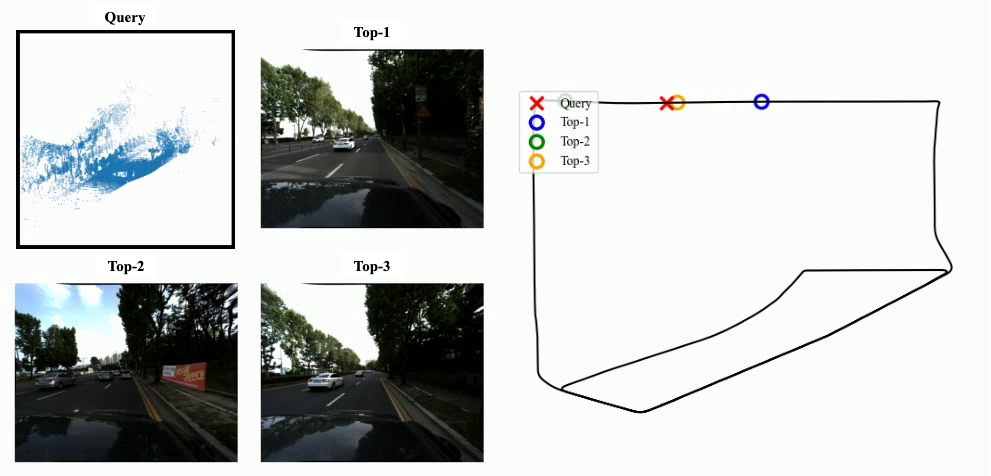}
    \caption{ViViD++ City day1-day2 3D-2D failed.}
  \vspace{-5pt}
    \label{fig:vividcity3d2dfail}
  \end{subfigure}
  \caption{Failure cases of 2D-3D and 3D-2D cross-modal retrieval with our VXP due to challenging and repetitive scenes from the ViViD++ benchmark. For each retrieval, the query and its top 3 retrievals are shown. Although the correct candidate is obtained within the top 3 places, the top 1 is not the closest in the latent embedding space, and thus, the performance is negatively affected.}
  \label{fig:failure_repeat}
  \vspace{-5pt}
\end{figure*}

Although VXP achieves state-of-the-art performance on cross-modal place recognition task and scores well in many challenging conditions, it fails in some cases. Primarily we have noticed that repetitive structures such as highway roads cause confusion for our method and lead to incorrect retrievals. Notably, uni-modal methods also fail in such cases as shown in \cref{fig:failure_repeat}. We believe that integrating sequential information as part of the mobile robot localization system would facilitate the task and remain part of future work.
\label{sec:fail_cases}

\begin{figure*}[t]
  \centering
  \begin{subfigure}{0.5\linewidth}
    \includegraphics[width=1\linewidth]{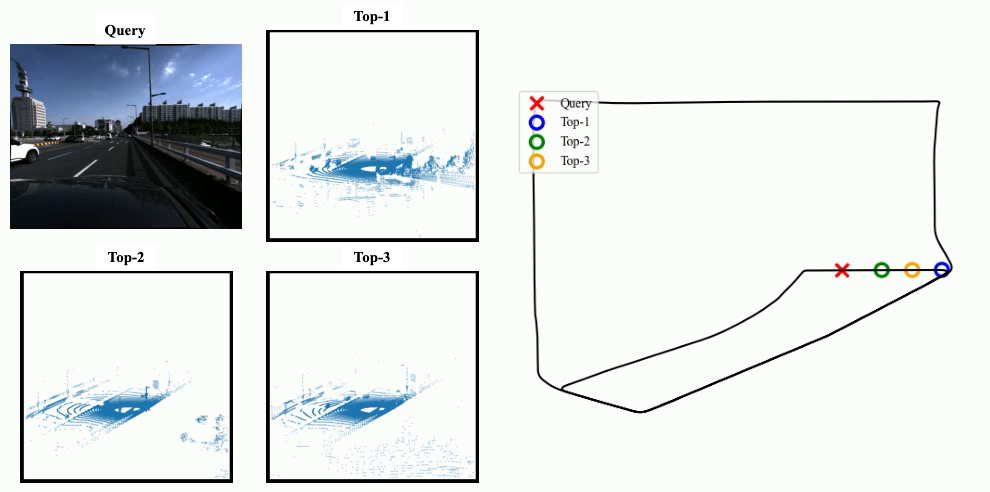}
    \caption{ViViD++ City day1-day2 2D-3D failed.}
  \vspace{-5pt}
    
    \label{fig:vividcityday122d3dfail}
  \end{subfigure}%
  \begin{subfigure}{0.5\linewidth}
    \includegraphics[width=1\linewidth]{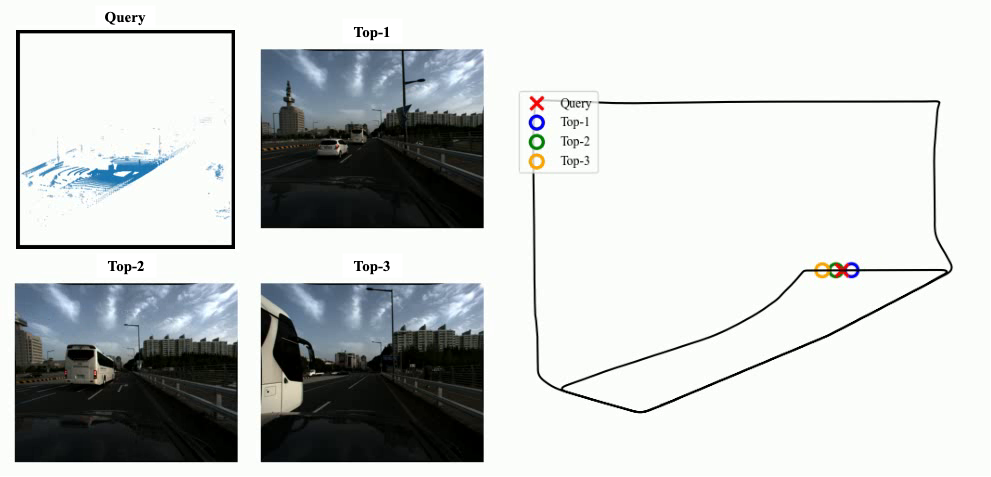}
    \caption{ViViD++ City day1-day2 3D-2D failed.}
  \vspace{-5pt}
    
    \label{fig:vividcityday123d2dfail}
  \end{subfigure}
  \caption{Failure cases of 2D-3D and 3D-2D cross-modal retrieval with our VXP due to sparse point cloud and lack of meaningful structures the ViViD++ benchmark. For each retrieval, the query and its top 3 retrievals are shown. Although the correct candidate is obtained within the top 3 places, the top 1 is not the closest in the latent embedding space, and thus, the performance is impaired.}
  \label{fig:failure_sparse}
\end{figure*}

\begin{figure*}[t]
  \centering
  \begin{subfigure}{0.5\linewidth}
    \includegraphics[width=1\linewidth]{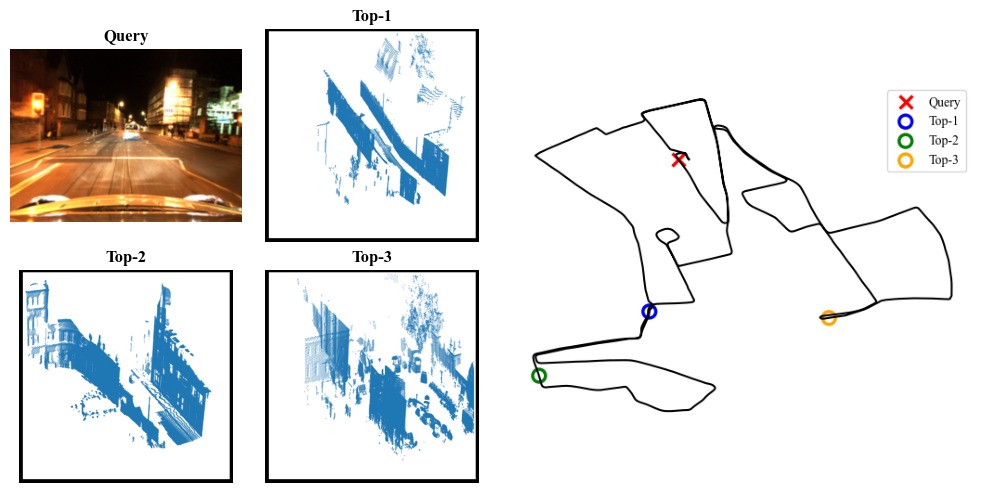}
    \caption{Oxford RobotCar night-overcast 2D-3D failed.}
    \label{fig:oxfordnightovercast2d3dfail}
  \end{subfigure}%
  \begin{subfigure}{0.5\linewidth}
    \includegraphics[width=1\linewidth]{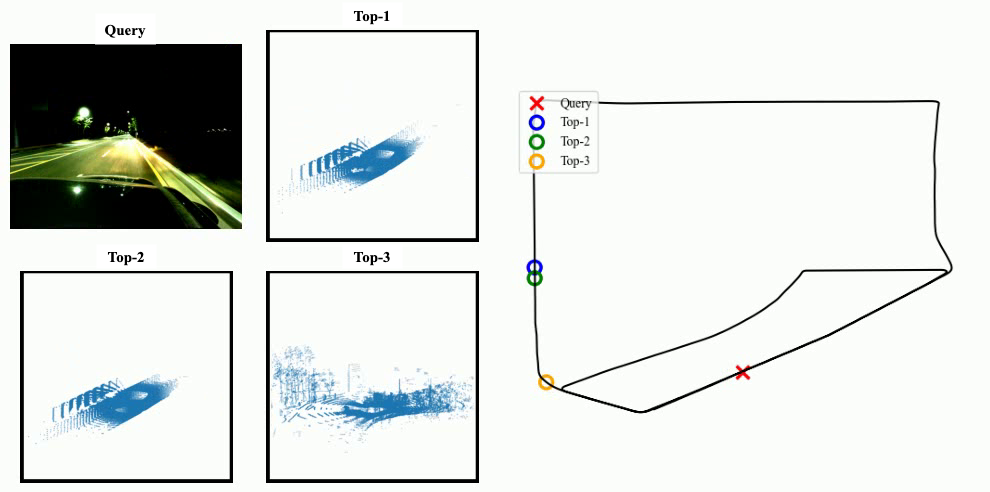}
    \caption{ViViD++ City night-day2 2D-3D failed.}
    \label{fig:vividcitynightday22d3dfail}
  \end{subfigure}
  \caption{Failure cases of 2D-3D cross-modal retrieval with our VXP due to poor illumination conditions in the night sequences from the Oxford RobotCar and ViViD++ datasets.}
  \label{fig:failure_light}
\end{figure*}

\noindent Secondly, sparse representation of places where the environment contains a lot of empty space and few, far-away structures, is not effective and lacks distinctive information. We demonstrate some failure examples in \cref{fig:failure_sparse}. One possible reason for worse performance lies in the projective nature of supervisory signal for our VXP. It is infeasible to learn a meaningful shared latent space and successfully perform cross-modal retrieval without establishing sufficient number of correspondences between voxels and pixels. Inspired by CASSPR~\cite{xia2023casspr}, it might be beneficial to incorporate a point branch and enhance point cloud encoding to tackle such challenging cases.\\
\noindent Lastly, illumination conditions are crucial in cross-modal retrieval where images are used as queries. Therefore, an image feature map generated from a poorly illuminated scene would obtain a bad-quality feature map, and leveraging such images for cross-modal retrievals becomes considerably challenging, as shown in \cref{fig:failure_light}. We believe that localizing with LiDAR scans (3D-2D retrieval) that are not affected by the light conditions would be more robust in such extreme cases.

\clearpage

\clearpage
{
    \small
    \bibliographystyle{ieeenat_fullname}
    \bibliography{main}
}

\end{document}